\documentclass{article}
\PassOptionsToPackage{numbers}{natbib}

\usepackage[final]{neurips_2021}

\usepackage[utf8]{inputenc} %
\usepackage[T1]{fontenc}    %
\usepackage{hyperref}       %
\usepackage{url}            %
\usepackage{booktabs}       %
\usepackage{amsfonts}       %
\usepackage{nicefrac}       %
\usepackage{microtype}      %
\usepackage[table]{xcolor}         %
\usepackage{xspace}
\usepackage{amsmath}
\usepackage{graphicx}
\usepackage{booktabs, multirow} %
\usepackage{soul}%
\usepackage{changepage,threeparttable} %
\usepackage[color=red]{todonotes}
\usepackage{lscape}
\usepackage{wrapfig}
\usepackage{caption}
\usepackage{afterpage}
\usepackage{enumitem}
\usepackage[linesnumbered,ruled,vlined]{algorithm2e}

\definecolor{links}{HTML}{2A1B81}
\hypersetup{colorlinks,linkcolor=,urlcolor=links}

\newcommand{\model}{FitVid\xspace}
\newcommand{\website}{\url{https://sites.google.com/view/fitvidpaper}\xspace}
\newcommand{\github}{\url{https://github.com/google-research/fitvid}\xspace}

\DeclareMathOperator{\E}{\mathbb{E}}

\newcommand{\stdgauss}{\mathcal{N}(\mathbf{0}, \mathbf{I})}

\newcommand{\bh}{\mathbf{h}}
\newcommand{\bx}{\mathbf{x}}
\newcommand{\bz}{\mathbf{z}}
\newcommand{\ba}{\mathbf{a}}
\newcommand{\bs}{\mathbf{s}}
\newcommand{\bxx}[2]{\bx_{#1:#2}}

\title{\model: Overfitting in Pixel-Level Video Prediction}

\author{%
  Mohammad Babaeizadeh\\
  Google Brain\\
  \texttt{mbz@google.com} \\
  \And
  Mohammad Taghi Saffar\\
  Google Brain\\
  \texttt{msaffar@google.com} \\
  \And
  Suraj Nair \\
  Stanford University \\
  \texttt{surajn@stanford.edu} \\
  \And
  Sergey Levine \\
  Google Brain \\
  \texttt{slevine@google.com} \\
  \And
  Chelsea Finn \\
  Google Brain \\
  \texttt{chelseaf@google.com} \\
  \And
  Dumitru Erhan \\
  Google Brain \\
  \texttt{dumitru@google.com} \\
}

\begin{document}
\maketitle
\vspace{-0.7cm}
\begin{abstract}
An agent that is capable of predicting what happens next can perform a variety of tasks through planning with no additional training. Furthermore, such an agent can internally represent the complex dynamics of the real-world and therefore can acquire a representation useful for a variety of visual perception tasks. This makes predicting the future frames of a video, conditioned on the observed past and potentially future actions, an interesting task which remains exceptionally challenging despite many recent advances. Existing video prediction models have shown promising results on simple narrow benchmarks but they generate low quality predictions on real-life datasets with more complicated dynamics or broader domain. There is a growing body of evidence that underfitting on the training data is one of the primary causes for the low quality predictions. In this paper, we argue that the inefficient use of parameters in the current video models is the main reason for underfitting. Therefore, we introduce a new architecture, named \model, which is capable of severe overfitting on the common benchmarks while having similar parameter count as the current state-of-the-art models. We analyze the consequences of overfitting, illustrating how it can produce unexpected outcomes such as generating high quality output by repeating the training data, and how it can be mitigated using existing image augmentation techniques. As a result, \model outperforms the current state-of-the-art models across four different video prediction benchmarks on four different metrics. 
\end{abstract}

\section{Introduction}
\label{sec:intro}

\begin{wrapfigure}[17]{r}{0.4\textwidth}
  \vspace{-1.6cm}
  \begin{center}
    \includegraphics[width=0.38\textwidth]{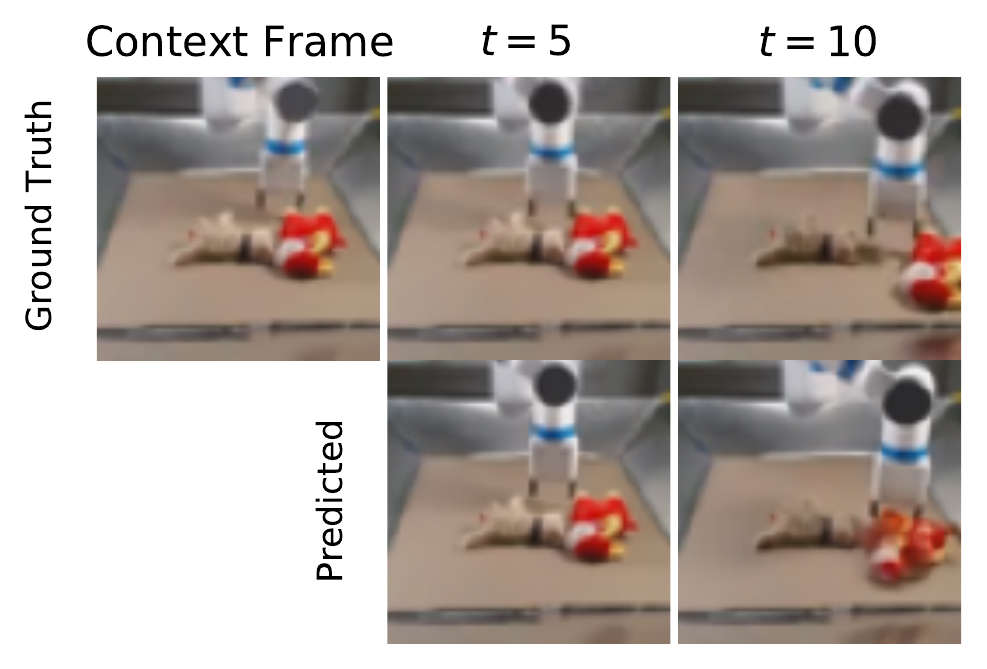}
  \end{center}
  \caption{\model is capable of predicting high quality images of the future given the first few frames. Note the accurately predicted movement of the pushed object with preserved visual details as well as the lack of movement from the stationary object. Also, note the detailed shadows of the robotic arm that are correctly predicted.}
\end{wrapfigure}

Predicting what happens next is a cornerstone of intelligence and one of the key capabilities of humans, which we heavily rely on to make decisions in everyday life~\cite{bubic2010prediction}. This capability enables us to anticipate future events and plan ahead to perform temporally extended tasks. While the machine learning literature has studied a wide range of prediction problems, one of the most direct challenges is to predict raw sensory inputs. In particular, prediction of future \emph{visual} inputs conditioned on a context of past observations -- i.e., pixel-level video prediction -- encapsulates the challenges of visual perception, modeling of physical events, and reasoning about uncertain behaviors. Video prediction can be formulated as a self-supervised problem, enabling us to use a substantial amount of unlabeled data to provide autonomous systems with powerful predictive capabilities as well as learning rich representations for downstream tasks. Already, video models have been successfully deployed in applications such as robotics~\cite{finn2017deep, zhang2019solar}, simulation~\cite{drivegan, kim2020learning}, compression~\cite{duan2020video, sulun2021can} and video synthesis from a single frame~\cite{endo2019animating, nam2019end}.

Despite recent advances in generative models in many domains, such as images~\cite{karras2017progressive, vahdat2020nvae} and text~\cite{devlin2018bert, brown2020language}, video prediction is still considered to be extremely challenging. The current state-of-the-art methods are limited to low-resolution videos (typically $64{\times}64$~\cite{villegas2019high} and a maximum of $256{\times}256$~\cite{drivegan}) usually in a narrow domain such as a single human walking, or a robotic arm pushing objects in a stationary setting. Even in such limited domains, the quality of predicted frames tend to drop substantially after less than 10 seconds into the future~\cite{saxena2021clockwork}. A growing body of evidence suggests that \textbf{underfitting} is one of the primary reasons for low quality predictions. For example,~\citet{villegas2019high} demonstrate how scaling the model, by simply adding more parameters, can substantially improve the prediction quality. Similarly,~\citet{castrejon2019improved} argue that blurry prediction of variational methods in video prediction is a sign of underfitting, exhibiting an improved test and train evidence lower bound as the network capacity increases.~\citet{wu2021greedy} also observed monotonic improvement as the number of modules in a hierarchical architecture increases. While scaling up models is a common trend in deep learning research to address underfitting, it comes at the cost of more computation, memory, and integration risks~\cite{wolf2019huggingface} as well as more complicated training regimes~\cite{wu2021greedy}. 

In this paper we take a step back and address underfitting by instead finding an architecture which uses its parameters more efficiently. More precisely, we propose \model, a model that -- with the same parameter count as current state-of-the-art models -- can significantly overfit to the video prediction datasets, including benchmarks that prior works have not been able to overfit to. To the best of our knowledge, this is the first time a video model reports \textit{substantial} overfitting on theses benchmarks. Importantly, we also find that simple image augmentation techniques can mitigate this overfitting, leading to models that can both fit the training set and generalize well to held-out videos. As a result, \model achieves state-of-the-art on four challenging video datasets across a wide range of metrics. Furthermore, we find that with \model we can utilize a significantly simpler training recipe. Prior works on video prediction, particularly those that make use of variational methods to provide for stochasticity, typically require a number of delicate design decisions to train successfully: curriculum training~\cite{oh2015action, finn2016unsupervised, lee2018stochastic, cenzato2019difficulty}, a learned prior~\cite{denton2018stochastic,villegas2019high,castrejon2019improved} and annealing of the weight on the VAE KL-divergence penalty~\cite{babaeizadeh2017stochastic}. In contrast to these approaches, we show that our method actually fits the training data well without any such components, training directly via optimizing the evidence lower bound with minimal hyperparameters. Videos generated by \model can be found in \website.

\vspace{-0.4cm}
\section{Related Work}
\vspace{-0.2cm}
Video prediction~\cite{ranzato2014video,srivastava2015unsupervised} has been formulated in different ways such as generating videos from a single image~\cite{endo2019animating, nam2019end, dorkenwald2021stochastic, yang2021dual, hu2021learning} or no image~\cite{vondrick2016generating,saito2018tganv2,clark2019adversarial}, text to video generation~\cite{godiva}, video-to-video translation~\cite{wang2019few, wang2018video} and data-driven simulation~\cite{kaiser2019model, drivegan, kim2020learning}. In this paper, our focus is on conditional video prediction, which is to predict the future frames of a video conditioned on a few initial context frames and possibly the future actions of the agents~\cite{finn2016unsupervised, babaeizadeh2017stochastic}. Conditional video prediction has a number of applications, including model-based reinforcement learning from pixels~\cite{hafner2018learning, hafner2019dream, kaiser2019model,rafailov2020offline} and
robotics~\cite{boots2014learning,finn2017deep,kalchbrenner2017video,ebert2017self,ebert2018robustness,ebert2018visual,xie2019improvisation,paxton2019visual,nair2019hierarchical,nair2020trass}.

Initially, video prediction was tackled using deterministic models~\cite{walker2015dense,finn2016unsupervised,jia2016dynamic,xue2016visual,walker2016uncertain,liang2017dual,byravan2017se3,vondrick2017generating,van2017transformation,liu2017video,chen2017video,lu2017flexible}. Later on, given the common randomness and partial observability in the real-life situations, various stochastic models were proposed to capture the stochasticity of the future. Generative adversarial networks~(GANs)~\cite{goodfellow2014generative} are demonstrated to generate sharp predictions~\cite{mathieu2015deep,lee2018stochastic,clark2019adversarial,luc2020transformation, hong2021arrowgan}; however, they tend to suffer from mode-collapse~\citep{goodfellow2016nips}, particularly in conditional generation settings~\citep{isola2016image}. Autoregressive video prediction models~\citep{kalchbrenner2016video,reed2017parallel,weissenborn2019scaling} can predict sharp but noisy videos while suffering from high training and inference time, particularly for longer videos. Flow based generative models~\cite{dinh2014nice, dinh2016density} are also adopted for video prediction~\cite{kumar2019videoflow} which can generate high quality videos but their high dimensional latent space makes them hard to implement and train. 

Variational auto-encoders~(VAEs)~\cite{kingma2013auto} are widely used for conditional video prediction as well~\cite{shu2016stochastic,babaeizadeh2017stochastic,denton2018stochastic,wichers2018hierarchical,villegas2019high,franceschi2020stochastic, castrejon2019improved,yan2021videogpt,rakhimov2020latent,lee2021revisiting,wu2021greedy,walker2021predicting}. Recently, it has been demonstrated that underfitting on training data plays a major role in low quality blurry predictions in VAE based models.~\citet{villegas2019high, castrejon2019improved} and \citet{wu2021greedy} all reported improved prediction quality as the network capacity increases. However, larger networks require more computation and memory. In this paper, we are interested in addressing underfitting \textit{without} increasing the capacity of the current models, and instead, find a more expressive architecture which uses its capacity more efficiently.

\section{Background}
Following prior work~\cite{finn2016unsupervised, babaeizadeh2017stochastic, denton2018stochastic, rubinstein1997optimization, oprea2020review, wu2021greedy}, we define the problem of pixel-level video prediction as follows: given the first $c$ frame of a video $\bx_{<c} = \bx_0, \bx_1, \ldots, \bx_{c-1}$, our goal is to predict the future frames by sampling from $p(\bxx{c}{T}|\bx_{<c})$. Optionally, the predictive model can be conditioned on additional given information $\ba_t$, such as the actions that the agents in the video are planning to take. This is typically called action-conditioned video prediction.

Variational video prediction follows the variational auto-encoder~\cite{kingma2013auto} formalism by introducing a set of latent variables $\bz$ to capture the inherent stochasticity of the problem. The latent variables can be fixed for the entire video~\cite{babaeizadeh2017stochastic} or vary over time~\cite{denton2018stochastic}. In both cases, we can factorize the likelihood model to $\prod_{t=c}^T p_\theta(\bx_t | \bx_{<t}, \bz_{\leq t})$ which is parametrized in an autoregressive manner over time; i.e. at each timestep $t$ the video frame $\bx_t$ and the latent variables $\bz_t$ are conditioned on the past latent samples and frames. By multiplying the prior we can factorize the predictive model as 
$$p(\bxx{c}{T}|\bx_{<c})=\prod\nolimits_{t=c}^T p_\theta(\bx_t | \bx_{<t}, \bz_{\leq t})p(\bz_t | \bx_{<t}, \bz_{<t})$$ 
where the prior $p(\bz)=p(\bz_t | \bx_{<t}, \bz_{<t})$ can be either fixed~\cite{kingma2013auto, babaeizadeh2017stochastic} or learned~\cite{chung2015recurrent, denton2018stochastic, castrejon2019improved}. For inference, we need to compute a marginalized distribution over the latent variables $\bz$, which is intractable. To overcome this problem, we use variation inference~\cite{jordan1999introduction} by defining an amortized approximate posterior $q(\bz|\bx)=\prod_{t} q(\bz_t|\bz_{<t}, \bx_{\leq t})$ that approximates the posterior distribution $p(\bz|\bx)$. The approximated posterior is commonly modeled with an inference network $q_\phi(\bz|\bx)$ that outputs the parameters of a conditionally Gaussian distribution $\mathcal{N}(\mu_\phi(\bx), \sigma_\phi(\bx))$. This network can be trained using the reparameterization trick~\citep{kingma2013auto}, according to:
$$\bz = \mu_\phi(\bx) + \sigma_\phi(\bx) \times \epsilon, \hspace{1cm} \epsilon \sim \mathcal{N}(\mathbf{0}, \mathbf{I})$$
Here, $\theta$ and $\phi$ are the parameters of the generative model and inference network, respectively. To learn these parameters, we can optimize the variational lower bound~\cite{kingma2013auto,rezende2014stochastic}:
\begin{equation}
\label{eqn:elbo}
\mathcal{L}(\bx) = -\E_{q_\phi(\bz|\bx)}\big[\log p_\theta(\bxx{t}{T}|\bx_{<t},\bz)\big] + \beta D_{KL}\big(q_\phi(\bz|\bx)||p(\bz)\big)    
\end{equation}

where $D_{KL}$ is the Kullback-Leibler divergence between the approximated posterior and the prior $p(\bz)$ which is fixed to $p(\bz)=\stdgauss$. The hyper-parameter $\beta$ represents the trade-off between minimizing frame prediction error and fitting the prior~\cite{higgins2016beta, higgins2016early,  denton2018stochastic}.

\begin{figure}[t]
  \centering
  \includegraphics[width=\textwidth]{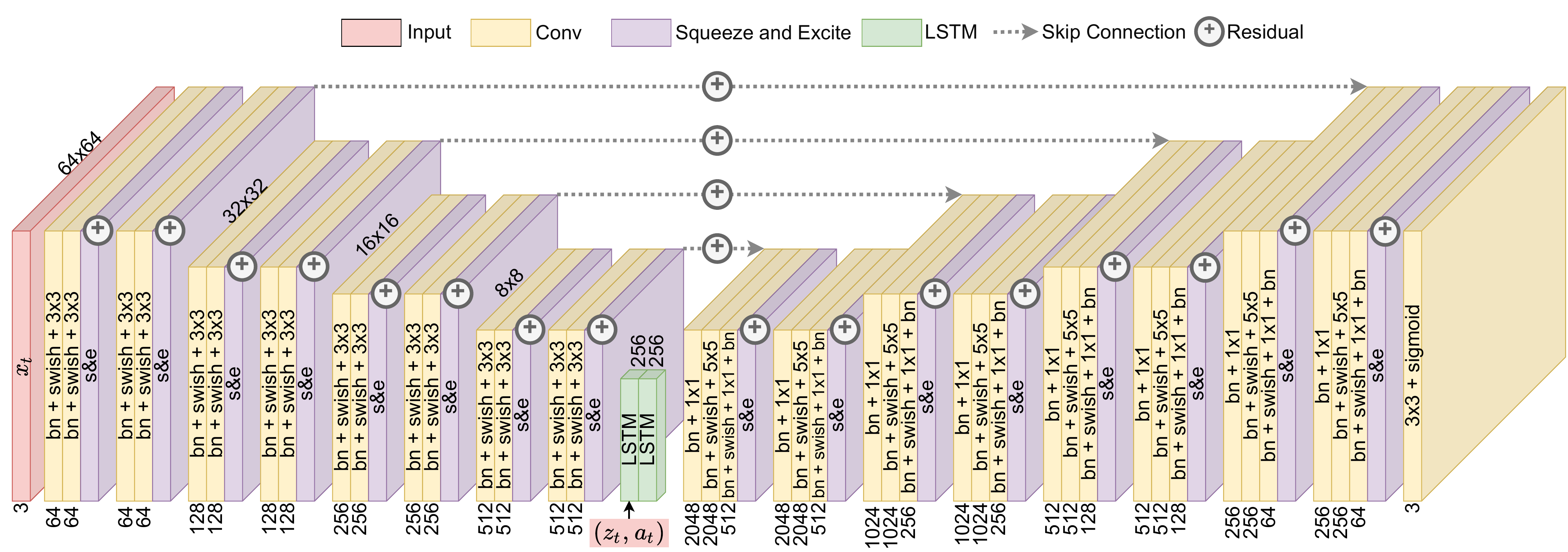}
  \caption{The \model architecture. In the figure, \textbf{(bn)} is batch-normalization~\cite{ioffe2015batch}, \textbf{(swish)} is the activation~\cite{ramachandran2017searching}, \textbf{(s\&e)} is Squeeze and Excite~\cite{hu2018squeeze} and $\mathbf{(N{\times}N)}$ is a convolutional layer with kernel size of $\mathbf{N{\times}N}$. The strides are always one, except when down-sampling which has a stride of two. For up-sampling we use nearest neighbour. The number under each box shows the number of filters while the top numbers indicate the input size. To model the dynamics, we use two layers of LSTMs~\cite{hochreiter1997long}.}
  \label{fig:model}
\end{figure}

\section{The \model Architecture} 
\label{sec:arc}
Striving for simplicity, we propose the \model model for stochastic video prediction, a convolutional non-hierarchical variational model with a fixed prior of $\stdgauss$. The architecture of \model is visualized in Figure~\ref{fig:model}.

\textbf{Encoder and decoder.} Following the recent advances in image generation, we use similar residual encoding and decoding cells as NVAE~\cite{vahdat2020nvae}. Each cell includes convolutional layers with batch-normalization~\cite{ioffe2015batch} and swish~\cite{ramachandran2017searching, elfwing2018sigmoid, hendrycks2016gaussian} as the activation function, followed by Squeeze and Excite~\cite{hu2018squeeze}. The encoder is made of four encoding blocks with two cells in each block. There is down-sampling after each encoder block using a strided convolution of size three in the spatial dimensions. The decoder also consists of four decoding blocks with two cells in each block, and a nearest neighbour up-sampling after each block. The number of filters in each encoding block is doubled while the number of filters in each decoding block is halved from the previous one. There is a residual skip connection between the encoder and the decoder after each cell which are fixed to the output from the last context frame. The statistics for batch-normalization is averaged across time.

\textbf{Dynamics model.} Similar to~\citet{denton2018stochastic}, the encoded frame $\bh_t$ is used to predict $\bh_{t+1}$ using two layers of LSTMs~\cite{hochreiter1997long}. Likewise, $q(\bz_t|\bx_{<t})$ is also modeled using a single layer of LSTMs with $\bh_{t+1}$ as the input that outputs the parameters of a conditionally Gaussian distribution $\mathcal{N}(\mu_\phi(\bx), \sigma_\phi(\bx))$. During the training, $\bz$ is sampled from $q(\bz_t|\bx_{<t})$ while at the inference time $\bz$ is sampled from the fixed prior $\stdgauss$. The input to the model is always the ground-truth image (which is usually referred to as teacher-forcing~\cite{goodfellow2016deep, cenzato2019difficulty}). At inference time, the predicted image in the previous time-step is used as input to predict the next frame. 

\textbf{Data augmentation.} We find that \model can \underline{substantially} overfit on some of the video prediction datasets~(read Section~\ref{sec:exp} and~\ref{sec:analysis}). To prevent the model from overfitting we use augmentation. To the best of our knowledge, this is the first use of augmentation in video prediction, perhaps because prior state-of-the-art models tend to underfit already and therefore would not benefit from it. Given the rich literature in image augmentation, we augment the videos using RandAugment~\cite{cubuk2020randaugment}. We randomize the augmentation per video but keep the randomization constant for frames of a single video. RandAugment substantially improves the overfitting, however not entirely, as it can be seen in Figure~\ref{fig:augment}. We improve the augmentation by selecting a random crop of the video before resizing it to the desired resolution at the training time, called RandCrop. The combination of RandCrop and RandAugment successfully prevents the overfitting, leading to models that both fit the training set and generalize well to held-out videos. 

\textbf{What \model does \textit{not} need.} Prior works on variational video prediction~\citep{finn2016deep, babaeizadeh2017stochastic, lee2018stochastic, villegas2019high, wu2021greedy}, generally require a range of additional design decisions for effective training. Common design parameters include using curriculum training, commonly by scheduled sampling~\cite{bengio2015scheduled}, to mitigate distributional shift between training and generation time~\citep{oh2015action, finn2016unsupervised, lee2018stochastic, cenzato2019difficulty}; heuristically tuning $\beta$ in Eqn~\ref{eqn:elbo} to balance the prediction vs fitting the prior~\cite{denton2018stochastic} by annealing it over the course of training~\cite{babaeizadeh2017stochastic, lee2018stochastic} or learned priors~\cite{denton2018stochastic,villegas2019high,castrejon2019improved,wu2021greedy}. Each of these design choices introduces hyperparameters, tuning burden, and additional work when applying a model to a new task. \model does not require any of these details: we simply train optimizing $\mathcal{L}(\bx)$ from Eqn~\ref{eqn:elbo} using Adam~\cite{kingma2014adam}.

Check the appendix for more architecture details. The source code is available at \\\github.

\section{Experiments}
\label{sec:exp}
To evaluate \model, we test it on four different real-world datasets and compare its performance with prior state-of-the-art methods, with comparable parameter count, using four different metrics. Our main goal is to demonstrate that \model can in fact overfit on these datasets and illustrate how augmentation can prevent \model from overfitting, resulting in state-of-the-art prediction performance. Please visit \website to see samples of videos.

\begin{table}[t]
\caption{The empirical comparison between \model (with 302M parameters), GHVAE~\cite{wu2021greedy} (with 599M parameters) and SVG~\cite{villegas2019high} (with 298M parameters). To prevent \model from overfitting, we use augmentation for Human3.6M and KITTI.
The green color highlights where \model achieved state-of-the-art result while the red color highlights otherwise.}
\label{tab:results}
\centering
\vspace{0.2cm}
\begin{minipage}{.48\textwidth}
    \centering
    \scriptsize
    \begin{tabular}{lrrrr}\toprule
    \textbf{RoboNet~\cite{dasari2019robonet}} &FVD$\downarrow$ &PSNR$\uparrow$ &SSIM$\uparrow$ &LPIPS$\downarrow$ \\\midrule
    GHVAE~\cite{wu2021greedy} &95.2 &24.7 &89.1 &0.036 \\
    SVG~\cite{villegas2019high} &123.2 &23.9 &87.8 &0.060 \\
    \model(ours) &\cellcolor[HTML]{d9ead3}\textbf{62.5} &\cellcolor[HTML]{d9ead3}\textbf{28.2} &\cellcolor[HTML]{d9ead3}\textbf{89.3} &\cellcolor[HTML]{d9ead3}\textbf{0.024} \\
    & & & & \\\toprule
    \textbf{KITTI~\cite{geiger2013vision}} &FVD$\downarrow$ &PSNR$\uparrow$ &SSIM$\uparrow$ &LPIPS$\downarrow$ \\\midrule
    GHVAE~\cite{wu2021greedy} &\cellcolor[HTML]{f4cccc}\textbf{552.9} &15.8 &\cellcolor[HTML]{f4cccc}\textbf{51.2} &0.286 \\
    SVG~\cite{villegas2019high} &1217.3 &15.0 &41.9 &0.327 \\
    \model(ours) &884.5 &\cellcolor[HTML]{d9ead3}\textbf{17.1} &49.1 &\cellcolor[HTML]{d9ead3}\textbf{0.217} \\
    \bottomrule
    \end{tabular}
\end{minipage}\hfill
\begin{minipage}{.48\textwidth}
    \centering
    \scriptsize
    \begin{tabular}{lrrrr}\toprule
    \textbf{Human3.6M~\cite{ionescu2014human3}} &FVD$\downarrow$ &PSNR$\uparrow$ &SSIM$\uparrow$ &LPIPS$\downarrow$ \\\midrule
    \multicolumn{5}{c}{Skip Frame of 1}\\\midrule
    GHVAE~\cite{wu2021greedy} &355.2 &26.7 &94.6 &0.018 \\
    SVG~\cite{villegas2019high} &- &- &- &0.060 \\
    \model(ours) &\cellcolor[HTML]{d9ead3}\textbf{154.7} &\cellcolor[HTML]{d9ead3}\textbf{36.2} &\cellcolor[HTML]{d9ead3}\textbf{97.9} &\cellcolor[HTML]{d9ead3}\textbf{0.012} \\\midrule
    \multicolumn{5}{c}{Skip Frame of 8}\\\midrule
    SVG~\cite{villegas2019high} &429.9 &23.8 &88.9 &- \\
    \model(ours) &\cellcolor[HTML]{d9ead3}\textbf{385.9} &\cellcolor[HTML]{d9ead3}\textbf{27.1} &\cellcolor[HTML]{d9ead3}\textbf{95.1} &0.026 \\
    \bottomrule
    \end{tabular}
\end{minipage}
\vspace{-0.5cm}
\end{table}

\subsection{Experimentation Setup}
\textbf{Datasets:}
To test \model, we use four datasets that cover a variety of real-life scenarios. We use the Human3.6M dataset~\cite{ionescu2014human3}, which consists of actors performing various actions in a room to study the structured motion prediction. We also use the KITTI dataset~\cite{geiger2013vision} to evaluate \model's ability to handle partial observability and dynamic backgrounds. For both datasets, we followed the pre-processing and testing format proposed by \citet{wu2021greedy} and \citet{villegas2019high}, which is to predict 25-frames conditioned the previous five in a $64{\times}64$ resolution.

To evaluate \model in an action-conditioned setting, we use RoboNet dataset~\citep{dasari2019robonet}. This large dataset includes more than 15 million video frames from 7 different robotic arms pushing objects in different bins. It contains a wide range of conditions, including different viewpoints, objects, tables, and lighting. Prior video prediction methods have a tendency to badly underfit on this dataset~\citep{dasari2019robonet}. Unfortunately, RoboNet does not provide a standard train/test partition. Hence, we follow the same setup as~\citet{wu2021greedy} and randomly select 256 videos for testing. Similar to~\citet{wu2021greedy}, we train \model to predict next ten frames given two context frames as well as the ten future actions. 

Finally, to compare \model to a wider range of prior work, we use the BAIR robot pushing dataset~\cite{2017arXiv171005268E}, which is a widely-used benchmark in the video prediction literature. We follow the evaluation protocol of~\citet{rakhimov2020latent}, which predicts the next 16 frames given only one context frame and no actions. Given the high stochasticity of robotic arm movement in BAIR, particularly in the action-free setting, it is a great benchmark for evaluating the model's ability to generate diverse outputs.

\textbf{Metrics:}
We evaluate our method and prior  models across four different metrics: Structural Similarity Index Measure~(SSIM)~\citep{wang2004image}, Peak Signal-to-noise Ratio~(PSNR)~\citep{huynh2008scope}, Learned Perceptual Image Patch Similarity~(LPIPS)~\cite{zhang2018perceptual} and Fréchet Video Distance~(FVD)~\cite{unterthiner2018towards}. FVD measures the overall visual quality and temporal coherence without reference to the ground truth video. PSNR, SSIM, and LPIPS measure pixel-wise similarity to the ground-truth with LPIPS most accurately representing human perceptual similarity. Given the stochastic nature of video prediction benchmarks, we follow the standard stochastic video prediction evaluation protocol~\cite{babaeizadeh2017stochastic, villegas2019high, wu2021greedy}: we sample 100 future trajectories per video and pick the best one as the final score for PSNR, SSIM and LPIPS. For FVD, we use all 100 with a batch size of 256.

\subsection{Results}
\textbf{Comparisons:}
First, we compare \model to GHVAE~\cite{wu2021greedy} and SVG~\cite{villegas2019high}. We chose these two baseline because they both investigated overfitting
by scaling the model, and achieve state-of-the-art results. However, SVG reported no overfitting even for their biggest model with 298M parameters~\cite{villegas2019high} while GHVAE (with 599M paramteres) reported "some overfitting" on smaller datasets~\cite{wu2021greedy}. At the same time, both of these models share a similar architecture to \model. GHVAE is a hierarchical variational video prediction model trained in a greedy manner. SVG is a large-scale variational video prediction model with learned prior and minimal inductive bias. As mentioned before, we compare against the largest version of SVG $(M=3, K=5)$ which has 298~million parameters that is in the same ballpark as \model with 302~million parameters.

\begin{figure}
  \centering
  \makebox[\textwidth][c]{\includegraphics[width=1.02\textwidth]{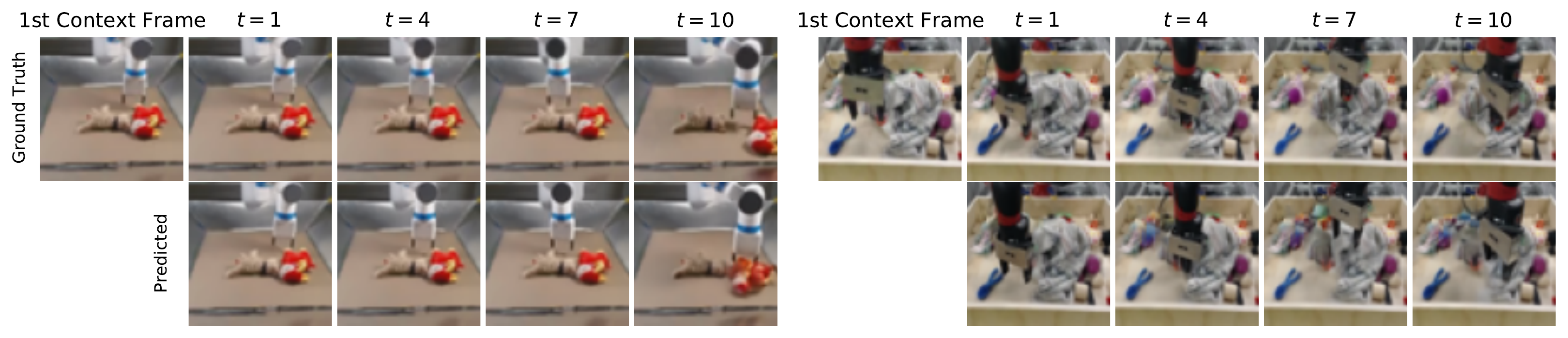}}
  \caption{\model on action-conditioned RoboNet~\cite{dasari2019robonet}. The model is conditioned on the first two frames and is predicting the next ten frames given the future actions of the robotic arm. These figures demonstrate how the predicted movements of the arm closely follows the ground truth given that the future actions is known. The model also predicts detailed movements of the pushed objects (visible in the left example) as well as filling in the previously unseen background with some random objects (look at the object that appear behind the robotic arm in the right). Also notice the wrong predictions of robot's fingers in the right example. See Figure~\ref{fig:robonet_full} for more frames from these video samples.}
  \label{fig:robonet}
  \vspace{0.3cm}

  \makebox[\textwidth][c]{\includegraphics[width=1.02\textwidth]{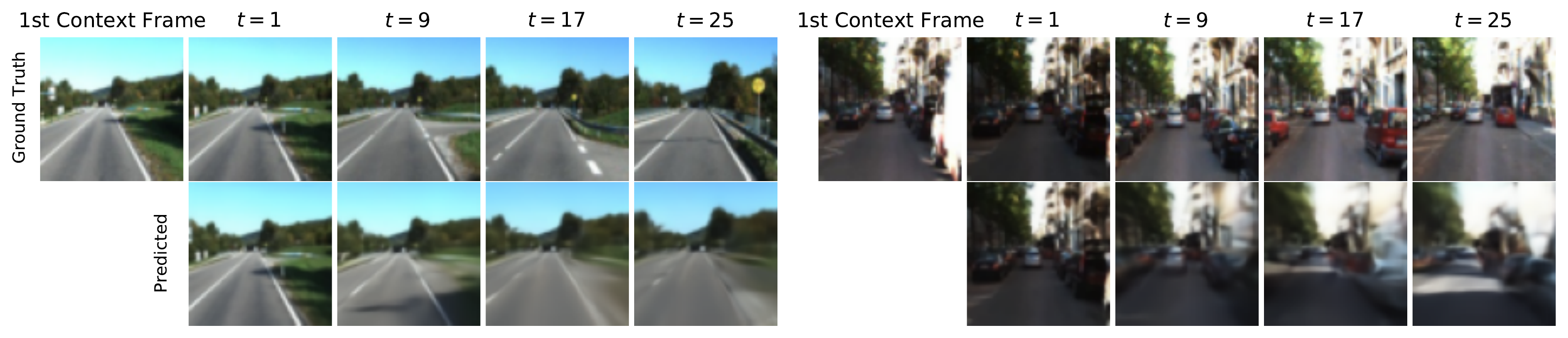}}
  \caption{\model on KITTI dataset~\cite{geiger2013vision}. As it can be seen in this figure, the model generates high quality prediction of the future in a dynamic scene. Note how in the top example \model keeps predicting the movement of the shadow on the ground till it gets out of the frame. After that, the model still brings the background closer in each frame, implying driving forward. We noticed that the quality of predictions drop substantially faster when there are more objects in the scene e.g. the driving scenes inside a city as can be seen in the right example. This indicates the model still fails to generalize to more complex scenes with more moving subjects. See Figure~\ref{fig:kitti_full} for more frames.}
  \label{fig:kitti}
  \vspace{0.3cm}
  
  \makebox[\textwidth][c]{\includegraphics[width=1.02\textwidth]{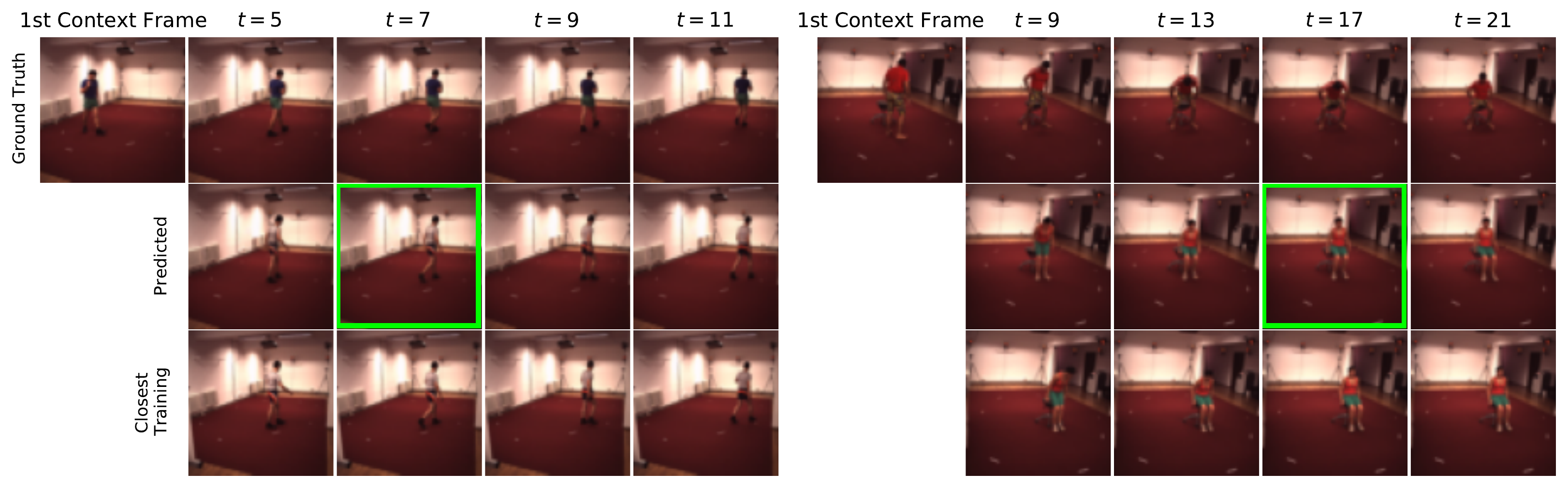}}
 
  \caption{\model on Human3.6M~\cite{ionescu2014human3}. This figures demonstrates extremely detailed and human-like motions predicted by \model, conditioned on the given context frames. However, on closer inspection, it can be seen that the human subject in the video is changing, from the test subject to a training subject. This is particularly evident from the cloths. This phenomena indicates that, although \model is capable of generalizing to the frames out of training distribution, however, it morphs the human subject into a familiar one from the training set and then \textbf{plays} the video from the \textbf{memory}. In fact, we can find similar videos in the training set as visualized in the last row. The highlighted frame is the one used for finding the closest training video. Check Figure~\ref{fig:humans_full} for more predicted frames.}
  \label{fig:humans}
  
  \centering{\bigskip\underline{More videos can be found at \website.}}

\end{figure}

Table~\ref{tab:results} contains the results of these experiments. As it can be seen in this table, \model outperforms both SVG and GHVAE across all metrics in Robonet and Human3.6M. In KITTI, \model also consistently outperforms SVG while either improves or closely matches the performance of GHVAE which has more than twice as parameters. For qualitative results, see Figures~\ref{fig:robonet},~\ref{fig:kitti} and \ref{fig:humans}.

\textbf{Comparison to non-variational methods:}
To compare the performance of \model with more prior methods, including non-variational models, we test it on BAIR robot pushing dataset~\cite{2017arXiv171005268E}. As can be seen in Table~\ref{tab:bair}, \model outperforms most of the previous models in this setting while performing comparably to Video Transformer~\cite{weissenborn2019scaling} which contains 373M parameters.

\begin{minipage}{.45\textwidth}
    \centering
    \includegraphics[width=\textwidth]{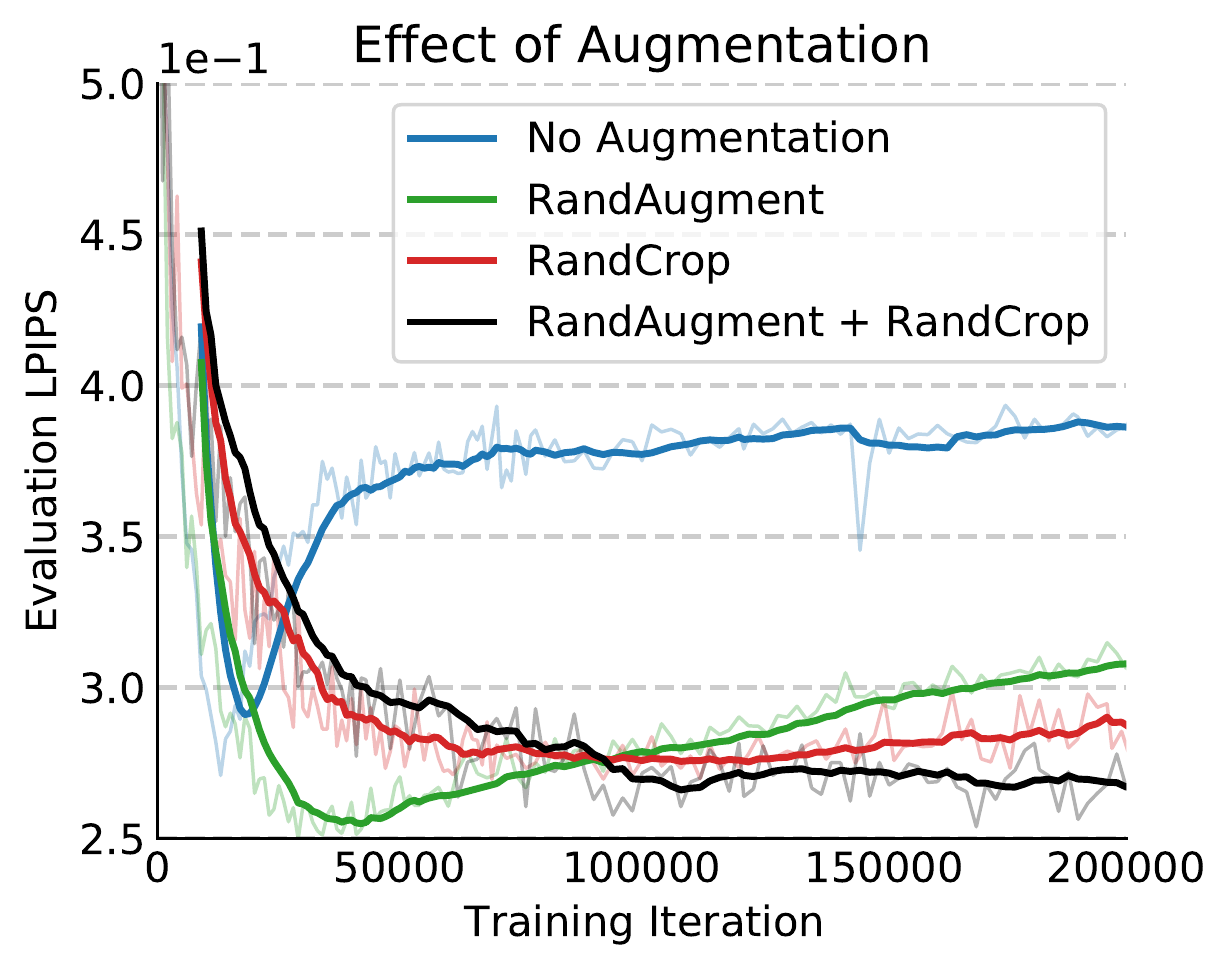}
    \captionof{figure}{Effect of augmentation methods on overfitting on KITTI. Both RandAugment and Random Crop improve the overfitting but the combination of two gets the best results.}
    \label{fig:augment}
\end{minipage}\hfill
\begin{minipage}{.45\textwidth}
    \centering
    \captionof{table}{Comparison between \model and different methods for video prediction on action-free BAIR dataset~\cite{2017arXiv171005268E}.}
    \label{tab:bair}
    \begin{tabular}{lrr}\toprule
    \textbf{BAIR} &FVD$\downarrow$ \\\midrule
    SV2P~\cite{babaeizadeh2017stochastic} &262.5 \\
    Latent Video Transformer~\cite{rakhimov2020latent} &125.8 \\
    SAVP~\cite{lee2018stochastic} &116.4 \\
    DVD-GAN-FP~\cite{clark2019adversarial} &109.8 \\
    VideoGPT~\cite{yan2021videogpt} &103.3 \\
    TrIVD-GAN-FP~\cite{luc2020transformation} &103.3 \\
    Video Transformer~\cite{weissenborn2019scaling} &94.0 \\
    \model(ours) &93.6 \\
    \bottomrule
    \end{tabular}
\end{minipage}

\section{Analysis}
\label{sec:analysis}

In this section, we take a closer look at the results from Section~\ref{sec:exp}, to analyse the consequences of overfitting and the effect of regularization on the current benchmarks.

\textbf{On Human3.6M as a video prediction benchmark:}
Human3.6M~\cite{ionescu2014human3} is a 
common benchmark in video prediction literature~\cite{finn2016unsupervised, babaeizadeh2017stochastic,villegas2017learning,wang2019memory,villegas2019high,lin2020motion,franceschi2020stochastic,guen2020disentangling,wu2021greedy,wu2021motionrnn} which we also use to evaluate \model (Figure~\ref{fig:humans}). At the first glance, it seems that the model is generating extremely detailed and human-like motions conditioned on the given context pose. However, on closer inspection, we observe that the human subject in the predicted video is changing. In fact, \model replaces the unseen human subject into a training subject which is particularly evident from the clothing. Actually, we can find similar video clips from the training data for each one of the predicted videos (see Figure~\ref{fig:humans}). These frames are \textbf{not} exactly the same, but they look notably similar. This observation indicates that:
\begin{enumerate}[leftmargin=0.55cm,noitemsep,topsep=0pt]
    \item The model can \textbf{generalize} to unseen frames and subjects since the test context frames are new and unseen. \model detects the human and continues the video from there.
    \item The model \textbf{memorized} the motion and the appearance of the training subjects. 
    The model \textit{morphs} the test human subject into a training one, and then \textit{plays} a relevant video from the memory. 
\end{enumerate}
This means that \model fails to generalize to a new subject, while still generalizing to unseen frames. Given that the Human3.6M has five training and two test subjects~\cite{finn2016unsupervised,villegas2017learning} this may not be surprising. 

Nevertheless, this observation shows how the current low-resolution setup for Human3.6M is not suitable for large-scale video prediction. In fact, after this observation, we traced the same behaviour in other video prediction literature and, unfortunately, it seems this is a common and overlooked issue. For example, the same phenomena can be seen in Figure~6 from \citet{franceschi2020stochastic} that shows changing the test to a training subject by Struct-VRNN~\cite{minderer2019unsupervised} and the proposed method by \citet{franceschi2020stochastic} (note the changed shirt). For others examples, see Figure~7 of \citet{villegas2019high} and Figure~5 of \citet{villegas2017learning}. A copy of these figures can be seen in Figure~\ref{fig:otherpapers} in the appendix.

\begin{figure}[t]
  \vspace{-0.5cm}
  \centering
  \includegraphics[width=\textwidth]{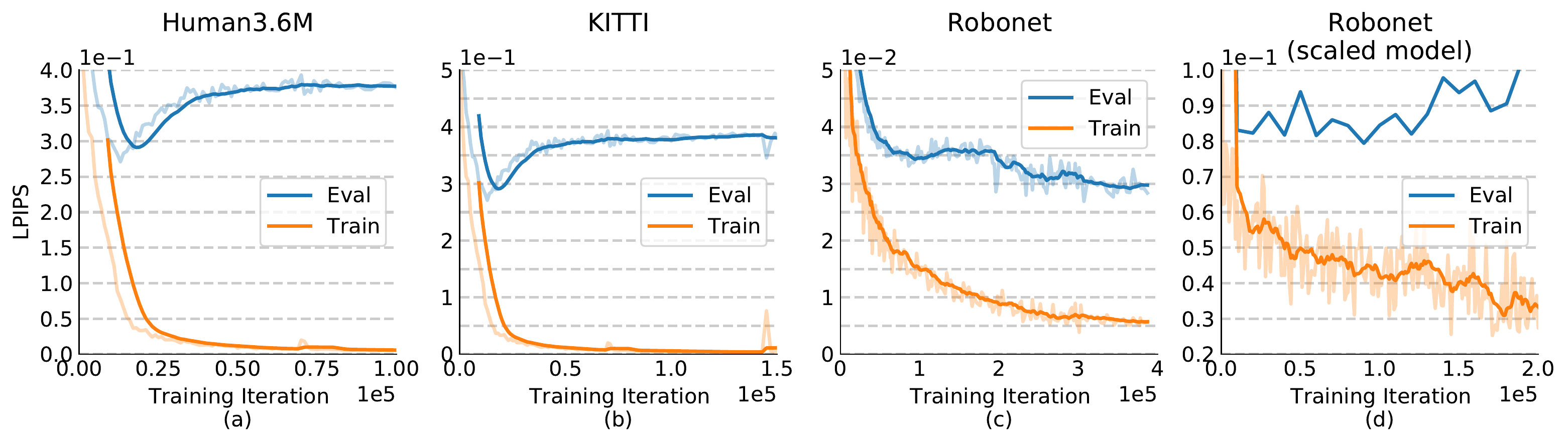}
  \caption{Overfitting of \model without augmentation. 
  This figure visualizes the training and evaluation metrics on \textbf{(a)}~Human3.6M~\cite{ionescu2014human3},  \textbf{(b)}~KITTI~\cite{geiger2013vision} and \textbf{(c)}~Robonet~\cite{dasari2019robonet}, without augmentation. As it can be seen, in all cases, \model overfits on the training data except for Robonet. This is evident from the evaluation measurement going up while the training keeps decreasing. In case of Robonet, \model with 302M parameters did \textbf{not} overfit but a scaled version of the model with 600M parameters did, as can be seen in \textbf{(d)}. $y$-axis is LPIPS. $x$-axis is the training iteration. The plots are smoothed with an average rolling window of size ten. The shadows are the raw non-smoothed values.}
  \label{fig:overfit}
  \vspace{-0.5cm}
\end{figure}

\textbf{Overfitting and regularization:}
As mentioned in Section~\ref{sec:intro}, there is considerable evidence that current video prediction models tend to underfit when trained on large datasets~\cite{villegas2019high, castrejon2019improved, wu2021greedy}. \citet{wu2021greedy}, which is the current state-of-the-art model with 599 million parameters, reported "some overfitting" on smaller datasets such as Human3.6M and KITTI. However, we observe severe and clear overfitting with \model, despite having only 302 million parameters. 
Figure~\ref{fig:overfit} visualizes the training and evaluation LPIPS metric while training \model on Human3.6M, \textbf{\underline{without augmentation}}. This graph demonstrates that the training keeps getting better while the test quality starts to get worse after $\sim$15K iterations. We also observed similar behaviour on KITTI, as can be seen in Figure~\ref{fig:overfit}b. These results clearly shows that \model is overfitting on Human3.6M and KITTI, indicating that \model is using its parameters more efficiently. As mentioned in Section~\ref{sec:arc}, to address overfitting, we use augmentation. As a result \model achieves state-of-the-art results as reported in Table~\ref{tab:results}.

\textbf{Overfitting on Robonet:}
We did not observe any overfitting on RoboNet, which is expected given the fact that RoboNet is much larger compared to the other benchmarks. Trying to find a model that can overfit on RoboNet, we test a scaled version of \model with 500M parameters -- which is still smaller compared to GHVAE with 599M parameters and reported no overfitting on this dataset. This scaled version of \model overfits on RoboNet, as demonstrated in Figure~\ref{fig:overfit}d. Note that we did \textbf{not} use this scaled version in the reported numbers of Table~\ref{tab:results}, which is generated using the 302M version. Our goal here was to demonstrate that a scaled version of \model can also use its parameters more efficiently, compared to prior models, leading to overfitting on even bigger datasets such as RoboNet.

\begin{wraptable}{r}{7cm}
\vspace{-0.2cm}
\caption{SVG~\cite{villegas2019high} with and without augmentation. This table shows that SVG does not benefit from augmentation as it is underfitting to the original data, as argued by \citet{villegas2017learning}. }
\label{tab:svgaug}
\centering
\vspace{0.1cm}
    \centering
    \scriptsize
    \begin{tabular}{lrrrr}\toprule
    \textbf{Human3.6M~\cite{ionescu2014human3}} &FVD$\downarrow$ &PSNR$\uparrow$ &SSIM$\uparrow$ &LPIPS$\downarrow$ \\\midrule
    Without &389.55 &27.4 &93.7 &0.041 \\
    With &429.25 &23.0 &87.1 &0.094 \\    
    & & & & \\\toprule
    \textbf{KITTI~\cite{geiger2013vision}} &FVD$\downarrow$ &PSNR$\uparrow$ &SSIM$\uparrow$ &LPIPS$\downarrow$ \\\midrule
    Without &1612.62 &14.8 &38.7 &0.330 \\
    With &2051.67 &14.4 &36.0 &0.333 \\
    \bottomrule
    \end{tabular}
\end{wraptable}

\textbf{Effect of Augmentation on SVG}
There is a discrepancy between the input data for training the models in Section~\ref{sec:exp}. \model is trained with augmentation while the  baselines are trained without any augmentation which raises a question: can the better performance of \model be explained only by the augmentation? In other words, do the previous methods benefit from augmentation too? To answer this question, we retrain SVG with and without augmentation. As demonstrated in Table~\ref{tab:svgaug}, SVG performs worse if trained with augmented data, supporting the claim that it is underfitting to the raw data. As a result, this experiment provides more support for \model truly overfitting on these datasets and therefore benefiting from augmentation. Please note that we included our SVG results without augmentation too, as we could not perfectly reproduce the numbers reported by ~\citet{villegas2019high} used in Table~\ref{tab:results}.

\begin{wraptable}[16]{r}{7cm}
\vspace{-0.2cm}
\caption{Zero-shot real robot performance. We use \model for planning future actions of a real robot pushing an object to a goal location with no training data from our setup. We train the model on visually different data (RoboNet) and the data from a closer domain (from ~\citet{wu2021greedy}) with and without augmentation. 
While unable to directly adapt from RoboNet to the new domain, the results illustrate that fine-tuning on similar data and augmentation improve \model's performance.
}
\label{tab:robot}
    \centering
    \scriptsize
    \begin{tabular}{lrr}\toprule
    Training Data & Success Rate\\\midrule
    Baseline (random actions) & 28\%\\
    RoboNet &17\%\\
    RoboNet + \citet{wu2021greedy} & 56\%\\
    RoboNet + Augmented \citet{wu2021greedy} & 78\%\\
    \bottomrule
    \end{tabular}
\end{wraptable}
\textbf{Zero-shot Real Robot Performance} Prior work indicate that improved video prediction translates to better performance in the downstream tasks~\cite{wu2021greedy,babaeizadeh2020models}. However, in these works, the training and test distribution are the same and there is almost no domain shift from training to testing. In this section, we are interested in investigating whether \model is capable of generalizing to a similar but visually different task with no training data for this new domain. Therefore, we setup a real-robot experiment, with a Franka Emika Panda robot arm, in which the goal is to push a specific object to a predetermined goal position. We train \model on RoboNet and use cross-entropy method~(CEM)~\citep{rubinstein1997optimization, chua2018deep} for planning (please see Appendix for details). As can be seen in Table~\ref{tab:robot}, this agent is unable to generalize to the new domain, achieving worse performance than a random agent. This may not be surprising given the fact that the videos in RoboNet have entirely different robots and visuals, although the robots are performing the same task (i.e. pushing objects in a bin using a robotic arm). We then try to bring the training and test domain closer to each other by fine-tuning \model on the data from ~\citet{wu2021greedy}. This data contains 5000 autonomously collected videos of a Franka Emika Panda robot arm pushing objects around which look more similar to our setup compared to RoboNet, but still contain different lighting, camera angle, and target objects. This time, we observe that \model is relatively successful at generalizing to the new domain, succeeding in $56\%$ of the trials. Finally, we find that adding data augmentation to the fine-tuning
improves the generalization ability of the model, achieving $78\%$ success rate. These results illustrate that while large distribution shift adaptation (RoboNet) remains difficult, by using data augmentation \model is capable of adapting to a relatively new domain (from \citet{wu2021greedy} data).

\section{Conclusion}
We propose \model, a simple and scalable variational video prediction model that can attain a significantly better fit to current video prediction datasets even with a similar parameter count as prior models. In fact, while prior methods generally suffer from \emph{underfitting} on these datasets, na\"{i}vely applying \model actually results in overfitting. We therefore propose a set of data augmentation techniques for video prediction that prevent overfitting, leading to state-of-the-art results across a range of prediction benchmarks.

To the best of our knowledge, this is the first time a model reports substantial overfitting on these benchmarks. This is particularly important because underfitting is usually cited as one the main reasons for low quality predictions of the future frames. We demonstrate how image augmentation techniques can prevent the model from overfitting, resulting in high quality images. As a result, \model outperformed the current state-of-the-art models across four different video prediction benchmarks on four different metrics. We also illustrate
how a model that can properly fit the training data, can fool the current benchmarks and metrics
resulting in undesired outcomes, which are often overlooked in the video prediction literature. 

There are many ways that \model can be expanded. As mentioned in the text, one of the interesting features of our proposed method is that it is simple. It is non-hierarchical, convolutional, with no attention mechanism, no curriculum learning, and no training scheduling. Any of these features can potentially improve the results of \model in order to generate even higher quality images. Given the simplicity of \model, it can be easily built upon. Another interesting direction would be to introduce new \textit{training-aware} metrics for video prediction and generation to signal when a model is generating high quality videos by repeating the training data. 

\section*{Broader Impact}
\label{sec:impact}
Videos are an abundant source of visual information about our physical world. They contain information about objects, humans and how they interact with each other. The goal of video prediction is to foremost learn a representation of the world, usable for downstream tasks by an agent. Second, is to predict what happens next, conditioned on the past and the future intents, which can be used for planning. Despite many recent advances in this field, the present day models are still relatively low-quality and limited to narrow domains which makes their applications limited. However, if improved, video prediction or representations learned by video prediction, can be a major step forward toward fully autonomous self-learning agents. This paper, we believe, takes an important step towards this goal by pushing state-of-the-art forward and simplifying it in a meaningful way. However, our model is vulnerable to the bias in the training data and if adopted widely, this can skew research in certain directions. For example, our results may lead to higher quality models which can be scaled to generate even higher quality results. These models will be harder to design and train and require more computational power, and potentially can be biased. Finally, the underlying techniques for video prediction can be misused for generating high-quality videos that are misleading, depicting deliberately false situations or persons, similar to the phenomenon of deepfakes~\citep{korshunov2018deepfakes}. However, we believe that our experiments are conducted in relatively specific narrow settings and conditions which will likely not generalize broadly. Specifically, it is especially challenging to generate high-quality realistic-looking videos even with such state of the art methods.

\small
\bibliographystyle{plainnat}
\bibliography{_references}

\clearpage
\appendix
\section{Appendix}

\subsection{Implementation Details}
\label{app:details}
In this section we describe the details of \model's architecture as well as training. Algorithm~\ref{alg:train} and Algorithm~\ref{alg:eval} describe the high-level training and prediction process respectively. Look at Table~\ref{tab:hyper} for the used hyper-parameters We used the same set of hyper-parameters across all experiments. As mentioned in Section~\ref{sec:arc}, all of the hyper-parameters are fixed during the training and there is no scheduling. Table~\ref{tab:encoder} and Table~\ref{tab:decoder} include the detailed architecture of the encoder and the decoder. Table~\ref{tab:poster} describes the structure of dynamic and posterior networks. Finally, Table~\ref{tab:augment} describes the augmentations details. 

\subsubsection{Computation Resources}
\label{app:resources}
We implement \model using Flax~\cite{flax2020github} library for JAX~\cite{jax2018github}. We train \model on $4{\times}4$ TPUs (32 co-processors). Each training step (with global batch size of 128 or local batch size of 4) takes $\sim1020$ milliseconds (i.e. $0.98$ step per second). In parallel, we evaluate the model every 1000 training steps on a single V100 GPU which takes about $\sim850$ milliseconds for a batch size of 128. 
The models are trained for one million training iterations which takes $\sim 12$ days to complete. 

\newcommand\mycommfont[1]{\scriptsize\ttfamily\textcolor{blue}{#1}}
\SetCommentSty{mycommfont}
\SetKwInput{KwInput}{Input}
\SetKwInput{KwOutput}{Output} 

\begin{minipage}[t]{0.46\textwidth}
\begin{algorithm}[H]
\footnotesize
\SetAlgoLined
\KwInput{Number of context frames $c$}
\KwData{Training frames $\bx_{0:T}$ and actions $\ba_{0:T}$}
    \tcp{Encode all frames}
    \For{$t\gets0$ \KwTo $T$}{
        $\bh_t,\mathbf{C}_t\gets$,  Encoder($\bx_t$)
    }

    \tcp{Prediction}
    $\bs_p, \bs_d\gets\mathbf{0},\mathbf{0}$\hspace{0.1cm}\tcp{Initialize states}
    \For{$t\gets0$ \KwTo $T$}{
        \tcp{Approximate posterior}
        $(\mathbf{\mathbf{\mu}}_t, \mathbf{\sigma}_t), \bs_p\gets$ Posterior([$\bh_{t+1}$], $\bs_p$)
        
        $\bz_t\sim\mathcal{N}(\mathbf{\mu}_t, \mathbf{\sigma}_t)$ 
        
        \tcp{Predict future state}
        $\hat{\bh}_t,\bs_d\gets$ Dynamic([$\bh_t$, $\ba_t$, $\bz_t$], $\bs_d$)
    }
    \tcp{Decode all frames}
    \For{$t\gets0$ \KwTo $T$}{
        \tcp{Use last available skip connection}
        $\hat{\bx}_t\gets$ Decoder($\hat{\bh}_t$, $\mathbf{C_c}$)
    }
    \tcp{Optimize ELBO}
    $\mathcal{L}\gets||\bx - \hat{\bx}||_2 + D_{KL}\big(\mathcal{N}(\mathbf{\mu}, \mathbf{\sigma}), \stdgauss \big)$
    
    $w\gets$Adam$\big(w, \mathcal{L}\big)$
 \caption{\model training.}
 \label{alg:train}
\end{algorithm}
\end{minipage}\hfill%
\begin{minipage}[t]{0.46\textwidth}
\begin{algorithm}[H]
\footnotesize
\SetAlgoLined
\KwInput{Context frames $\bx_{0:c}$}
\KwInput{All actions $\ba_{0:T}$}
\KwOutput{$\hat{\bx}$}
    $\bs_d\gets\mathbf{0}$\hspace{0.1cm}\tcp{Initialize states}
    $_, \mathbf{C}\gets$Encoder($\bx_c$) \tcp{Get last skips.}
    \For{$t\gets0$ \KwTo $T$}{
        \tcp{Encode frame}
        $\bh_t,\mathbf{C}_t\gets$,  Encoder($\bx_t$)
        
        \tcp{Sample from prior}
        $\bz_t\sim\mathcal{N}(\stdgauss)$ 
        
        \tcp{Predict future state}
        $\hat{\bh}_t,\bs_d\gets$ Dynamic([$\bh_t$, $\ba_t$, $\bz_t$], $\bs_d$)
        
        \tcp{Decode frame}
        $\hat{\bx}_t\gets$ Decoder($\hat{\bh}_t$, $\mathbf{C_c}$)
    }
 \caption{\model prediction.}
 \label{alg:eval}
\end{algorithm}
\end{minipage}

\begin{table}[!htp]\centering
\caption{Hyper-parameters used for training \model. We used the same set of hyper-parameters across all experiments. As mentioned in Section~\ref{sec:arc}, all of the hyper-paramters are fixed during the training and there is no scheduling.}
\label{tab:hyper}
\scriptsize
\begin{tabular}{lr}\toprule
Hyper-parameter &Value \\\midrule\midrule
\multicolumn{2}{c}{Optimizer (Adam\cite{kingma2014adam})} \\
\midrule
Learning Rate ($\alpha$) &$1\mathrm{e}{-3}$ \\
Btach Size &$128$ \\
$\beta_1$ &$0.9$ \\
$\beta_2$ &$0.999$ \\
$\epsilon$ &$1\mathrm{e}{-8}$ \\
Gradient Clipping ($l_2$) &100.0 \\
\midrule
\multicolumn{2}{c}{Model} \\
\midrule
$\beta$ &1  \\
Latent~($\bz$)-dimension &10 \\
Encoder~($\bh$) dimension &128 \\
LSTM size &256 \\
\bottomrule
\end{tabular}
\end{table}

\begin{table}[!htp]\centering
\caption{\model Encoder Architecture. We are using the same encoding cells as NVAE~\cite{vahdat2020nvae}. The strides are always $1{\times}1$ except when down-sampling which has strides of $2{\times}2$. \textbf{(bn)} is batch-normalization~\cite{ioffe2015batch}, \textbf{(swish)} is the activation~\cite{ramachandran2017searching}, \textbf{(s\&e)} is Squeeze and Excite~\cite{hu2018squeeze}. There is a skip connection from the beginning of each cell to the end of it. In these skip connections, the number of input filters will be matched by the output using a $1{\times}1$ convolution.}
\label{tab:encoder}
\scriptsize
\begin{tabular}{lccccccc}\toprule
Cell &Input Size &Pre &Kernel &Filters &Post &Down Sampling \\\midrule
1-1 &$64{\times}64$ &bn + swish &$3{\times}3$ &64 &- &- \\
1-1 &$64{\times}64$ &bn + swish &$3{\times}3$ &64 &s\&e &- \\
1-2 &$64{\times}64$ &bn + swish &$3{\times}3$ &64 &- &- \\
1-2 &$64{\times}64$ &bn + swish &$3{\times}3$ &64 &s\&e &Yes \\
2-1 &$32{\times}32$ &bn + swish &$3{\times}3$ &128 &- &- \\
2-1 &$32{\times}32$ &bn + swish &$3{\times}3$ &128 &s\&e &- \\
2-2 &$32{\times}32$ &bn + swish &$3{\times}3$ &128 &- &- \\
2-2 &$32{\times}32$ &bn + swish &$3{\times}3$ &128 &s\&e &Yes \\
3-1 &$16{\times}16$ &bn + swish &$3{\times}3$ &256 &- &- \\
3-1 &$16{\times}16$ &bn + swish &$3{\times}3$ &256 &s\&e &- \\
3-2 &$16{\times}16$ &bn + swish &$3{\times}3$ &256 &- &- \\
3-2 &$16{\times}16$ &bn + swish &$3{\times}3$ &256 &s\&e &Yes \\
4-1 &$8{\times}8$ &bn + swish &$3{\times}3$ &512 &- &- \\
4-1 &$8{\times}8$ &bn + swish &$3{\times}3$ &512 &s\&e &- \\
4-2 &$8{\times}8$ &bn + swish &$3{\times}3$ &512 &- &- \\
4-2 &$8{\times}8$ &bn + swish &$3{\times}3$ &512 &s\&e &- \\
\bottomrule
\end{tabular}
\end{table}

\begin{table}[!htp]\centering
\caption{\model Dynamics Architecture. We are using a similar dynamics as~\citet{denton2018stochastic}. The encoded output is first averaged across spatial dimension and then decoded into $\bh$-size using a fully connected layer. Then, the dynamics are modeled by two LSTM layers. Finally, the output is mapped and reshaped to an image tensor before passing to the decoder. The posterior uses the exact same architecture except that only has one LSTM layer.}
\label{tab:poster}
\scriptsize
\begin{tabular}{cccccc}\toprule
Cell &Input Size &Pre &Layer &Size &Post\\\midrule
- &$8{\times}8$ &spatial average + flatten &dense &256 &append $\bz$ and $\ba$ \\
- &128+ & - &LSTM &256 &- \\
- &256 & - &LSTM &256 &- \\
- &256 & - &dense &$8{\times}8{\times}512$ &sigmoid + reshape \\
\bottomrule
\end{tabular}
\end{table}

\begin{table}[!htp]\centering
\caption{\model Encoder Architecture. We are using the same encoding and decoding cells as NVAE~\cite{vahdat2020nvae}. The strides are always $1{\times}1$. For up-sampling we use nearest neighbour. \textbf{(bn)} is batch-normalization~\cite{ioffe2015batch}, \textbf{(swish)} is the activation~\cite{ramachandran2017searching}, \textbf{(s\&e)} is Squeeze and Excite~\cite{hu2018squeeze}. There is a skip connection from the beginning of each cell to the end of it. There are also skip connections from each encoder block to the corresponding decoder block (look at Figure~\ref{fig:model}). In these skip connections, the number of input filters will be matched by the output using a $1{\times}1$ convolution.}
\label{tab:decoder}
\scriptsize
\begin{tabular}{ccccccc}\toprule
Cell &Input Size &Pre &Kernel &Filters &Post &Up Sampling \\\midrule
1-1 &$8{\times}8$ &bn &$1{\times}1$ &2048 &- &- \\
1-1 &$8{\times}8$ &bn + swish &$5{\times}5$ &2048 &- &- \\
1-1 &$8{\times}8$ &bn + swish &$1{\times}1$ &512 &bn + s\&e &- \\
1-2 &$8{\times}8$ &bn &$1{\times}1$ &2048 &- &- \\
1-2 &$8{\times}8$ &bn + swish &$5{\times}5$ &2048 &- &- \\
1-2 &$8{\times}8$ &bn + swish &$1{\times}1$ &512 &bn + s\&e &Yes \\
2-1 &$16{\times}16$ &bn &$1{\times}1$ &1024 &- &- \\
2-1 &$16{\times}16$ &bn + swish &$5{\times}5$ &1024 &- &- \\
2-1 &$16{\times}16$ &bn + swish &$1{\times}1$ &256 &bn + s\&e &- \\
2-2 &$16{\times}16$ &bn &$1{\times}1$ &1024 &- &- \\
2-2 &$16{\times}16$ &bn + swish &$5{\times}5$ &1024 &- &- \\
2-2 &$16{\times}16$ &bn + swish &$1{\times}1$ &256 &bn + s\&e &Yes \\
3-1 &$32{\times}32$ &bn &$1{\times}1$ &512 &- &- \\
3-1 &$32{\times}32$ &bn + swish &$5{\times}5$ &512 &- &- \\
3-1 &$32{\times}32$ &bn + swish &$1{\times}1$ &128 &bn + s\&e &- \\
3-2 &$32{\times}32$ &bn &$1{\times}1$ &512 &- &- \\
3-2 &$32{\times}32$ &bn + swish &$5{\times}5$ &512 &- &- \\
3-2 &$32{\times}32$ &bn + swish &$1{\times}1$ &128 &bn + s\&e &Yes \\
4-1 &$64{\times}64$ &bn &$1{\times}1$ &256 &- &- \\
4-1 &$64{\times}64$ &bn + swish &$5{\times}5$ &256 &- &- \\
4-1 &$64{\times}64$ &bn + swish &$1{\times}1$ &64 &bn + s\&e &- \\
4-2 &$64{\times}64$ &bn &$1{\times}1$ &256 &- &- \\
4-2 &$64{\times}64$ &bn + swish &$5{\times}5$ &256 &- &- \\
4-2 &$64{\times}64$ &bn + swish &$1{\times}1$ &64 &bn + s\&e &- \\
-&$64{\times}64$ &- &$1{\times}1$ &3 &sigmoid &- \\
\bottomrule
\end{tabular}
\end{table}

\begin{table}
\caption{To prevent \model from overfitting we use augmentation. First, at training time, we select a random crop of the video before resizing it to the desired resolution ($64{\times}64$) at the training time, called RandCrop. This processes crops all the frames of a given video to include a minimum of $C$ percent of the frame's height. Then we use RandAugment~\cite{cubuk2020randaugment} to improve the augmentation. We use the same augmentation configuration for all the datasets. Per video, we use the same randomization across all the frames.}
\label{tab:augment}
\begin{minipage}{0.52\textwidth}
    \begin{algorithm}[H]
    \footnotesize
    \SetAlgoLined
    \KwInput{Video $\bx$}
    \KwInput{Number of RandAugment transformations $N$}
    \KwInput{RandAugment magnitude $M$}
    \KwInput{RandCrop crop height minimum ratio $C$}

    $\bx\gets RandCrop(\bx, C)$

        \For{$i\gets0$ \KwTo $N$}{
            $f\gets$ ChooseRandomTransformation()
            
            $\bx\gets f(\bx, M)$
        }
    \Return{$\bx$}
    
     \caption{Video Augmentation.}
     \label{alg:aug}
    \end{algorithm}
\end{minipage}\hfill
\begin{minipage}{0.36\textwidth}
    \scriptsize
    \begin{tabular}{lr}\toprule
    Hyper-parameter &Value \\\midrule\midrule
    \multicolumn{2}{c}{RandAugment} \\
    \midrule
    No. of transformations $N$ &1  \\
    Magnitude $M$ &5 \\ \midrule
    \multicolumn{2}{c}{Transformations} \\
    \midrule
    identity - auto\_contrast - equalize & \\
    rotate - solarize - color - posterize & \\
    contrast - brightness - sharpness& \\
    shear\_x - shear\_y & \\
    translate\_x - translate\_y & \\
    \\ \midrule
    \multicolumn{2}{c}{RandCrop} \\\midrule
    \midrule
    Crop height minimum ratio $C$ &$0.8$ \\
    \bottomrule
    \end{tabular}
\end{minipage}
\end{table}

\begin{table}[]
    \caption{Used datasets and their licenses.}
    \vspace{0.5cm}
    \centering
    \begin{tabular}{lll}
         Dataset & Reference & License \\ \midrule
         RoboNet & \citet{dasari2019robonet} & \href{https://github.com/SudeepDasari/RoboNet/blob/master/LICENSE}{MIT}\\
         KITTI & \citet{Geiger2013IJRR} & \href{http://www.cvlibs.net/datasets/kitti/}{Creative Commons} \\
         Human3.6M & \citet{ionescu2014human3} & \href{http://vision.imar.ro/human3.6m/eula.php}{License} \\
         BAIR robot pushing & \citet{ebert2017self} & \href{https://docs.github.com/en/github/creating-cloning-and-archiving-repositories/creating-a-repository-on-github/licensing-a-repository}{GitHub Default} \\    \end{tabular}
    \label{tab:dataset}
\end{table}

\begin{figure}
  \centering
  \includegraphics[width=\textwidth]{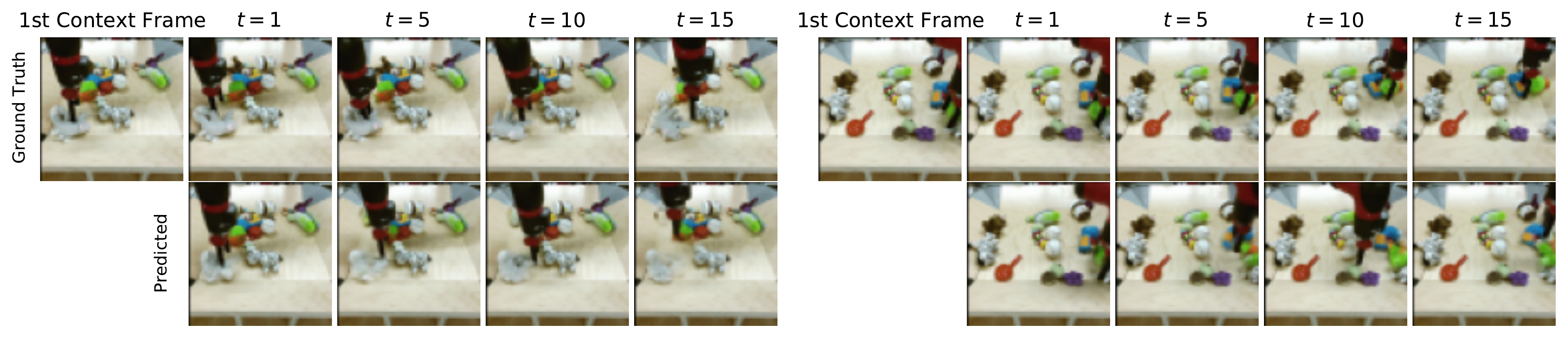}
  \caption{\model on BAIR robot pushing dataset~\cite{2017arXiv171005268E} with no actions. The model is conditioned only on the first frame and is predicting the next 16 frames. Given that the future actions of the robotic arm is unknown, the prediction can diverge substantially from the ground truth video. However, the model predicts movements for the objects whenever the arm pushes the object in an imaginary scenario. It also fills the background with random objects.}
  \label{fig:bair}
\end{figure}

\subsection{Robot Experiment Details}
\label{app:robot}

\subsubsection{Environment}

The robot environment consists of a Franka Emika Panda robot operating over a bin which contains various objects. The robot's observations are $64{\times}64{\times}3$ RGB images, and its action space consists of 3 DOF delta position control of the end effector with action magnitudes in the range [-10cm,10cm]. 

\subsubsection{Data}

The data used for finetuning the model used in robot experiments was taken directly from \citet{wu2021greedy}. This data consists of different viewpoint, lighting conditions, and target objects than what is used in our evaluation.

\subsubsection{Evaluation Details}

\begin{figure}[t]
  \centering
  \includegraphics[width=\textwidth]{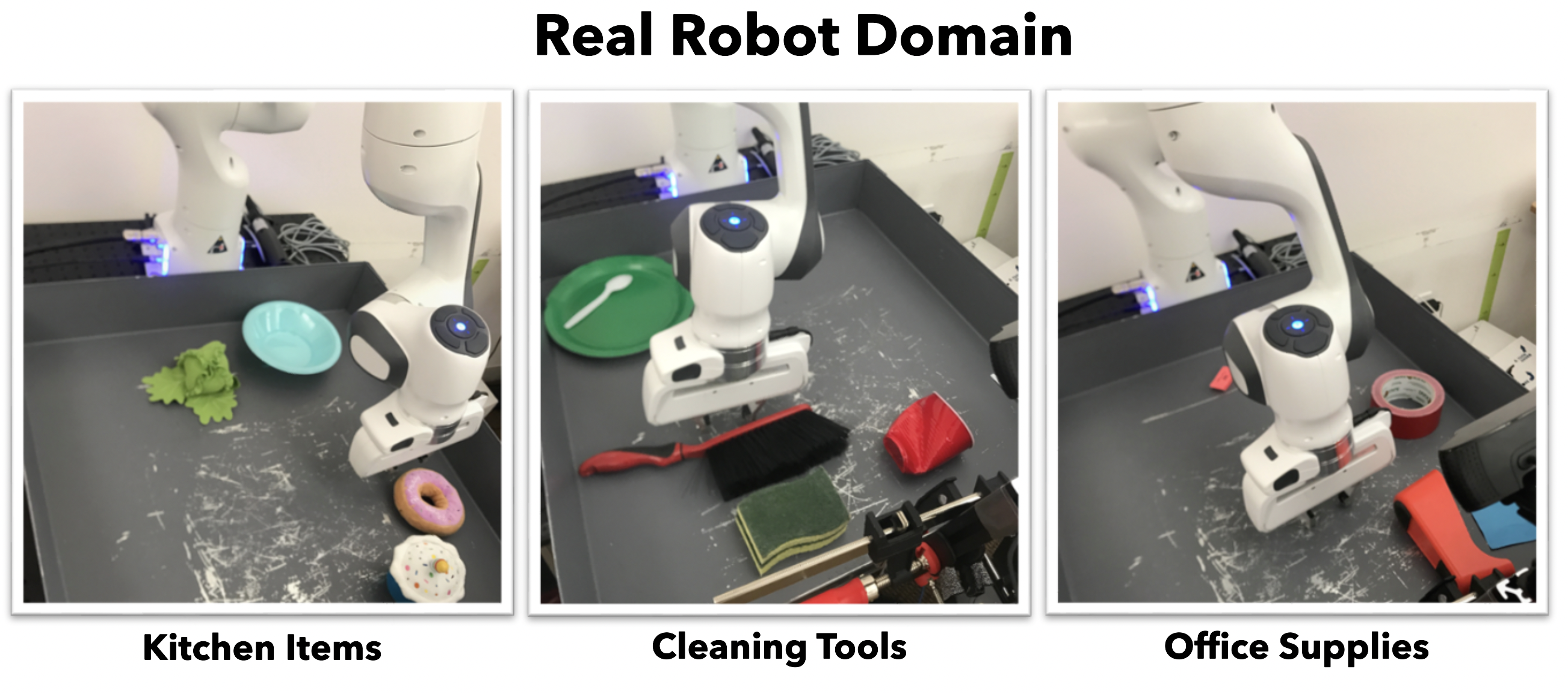}
  \caption{Zero-shot robot domain. We evaluate \model on planning tasks on a Franka Emika Panda involving kitchen, cleaning, and office items. We did not collect any training data for this task. Instead the model is trained on RoboNet and fine-tuned on augmented data from~\citet{wu2021greedy}.}
  \label{fig:robotdomains}
\end{figure}

During evaluation, the agent is specified to complete an object pushing task by a goal image. The agent has 50 timesteps to complete the task, and a trial is measured as successful if the majority of the object overlaps with its target position at some point in the episode. Each method is evaluated over 18 trials, of which 6 consist of objects in an ``office'' setting, 6 consist of objects in a ``kitchen'' setting, and 6 consist of objects in a ``cleaning'' setting (See Figure~\ref{fig:robotdomains}).

To execute the task, the agent performs visual model predictive control using the cross entropy method (CEM). Specifically, the agent takes in 1 frame, and predicts trajectories of 10 time-steps for 200 different sampled action sequences. Trajectories are ranked according to their negative mean squared error to the goal image, averaged across all 10 time-steps. The action distribution refits to the top 20 actions, and repeats for 3 iterations of CEM. Afterwards the best sequence of 10 actions is stepped in the environment in an open loop faction. The process repeats 5 times until the end of the episode.

\begin{figure}[b]
  \centering
  \includegraphics[width=\textwidth]{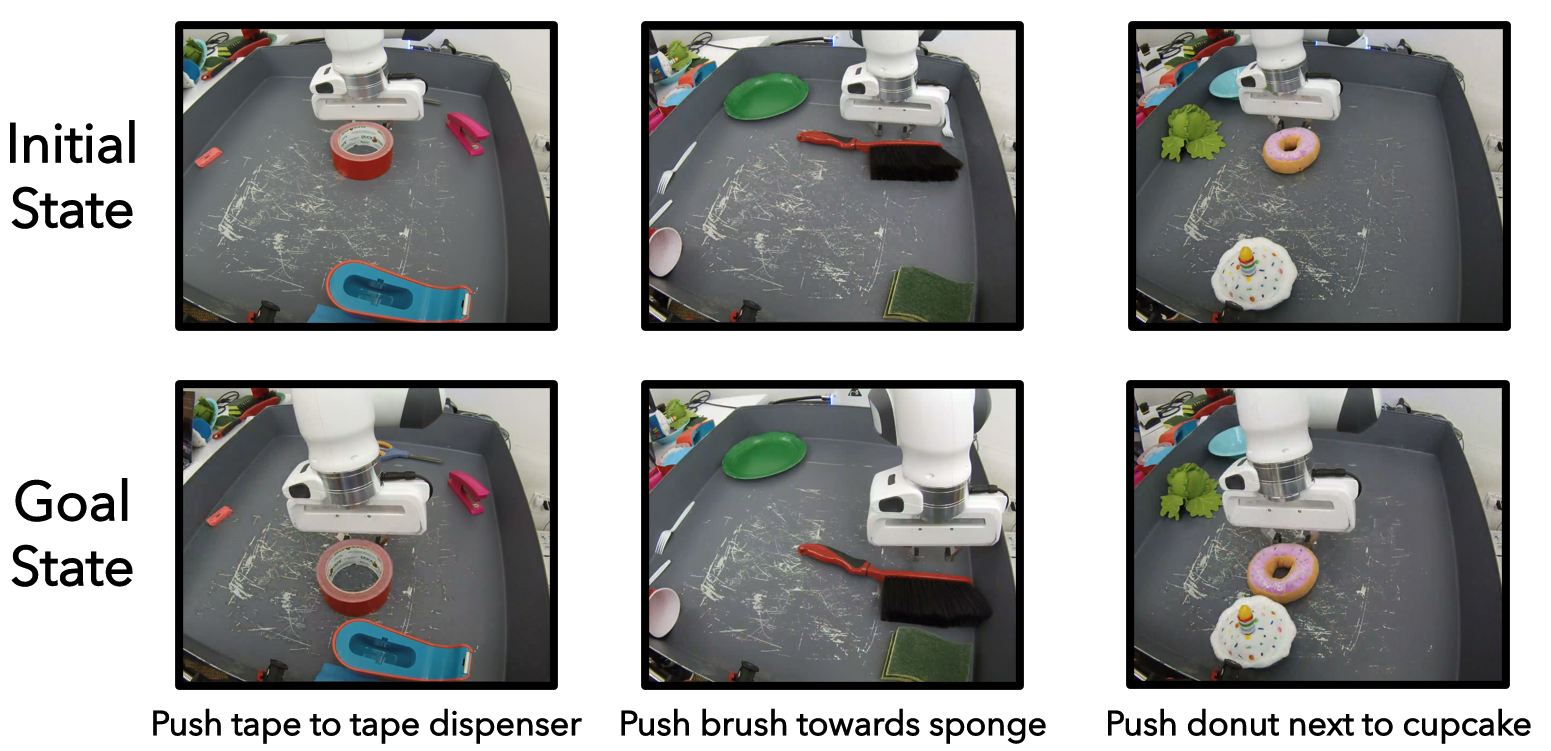}
  \caption{Example tasks for zero-shot object pushing using a robotic arm. The goal in each trial is to push the a specific object to a predetermined goal location. The trial is considered successful, if the robot pushes at least half of the object overlaps with its goal location at any point in the episode.}
\end{figure}

\begin{landscape}
    \begin{figure}
        \includegraphics[width=\linewidth]{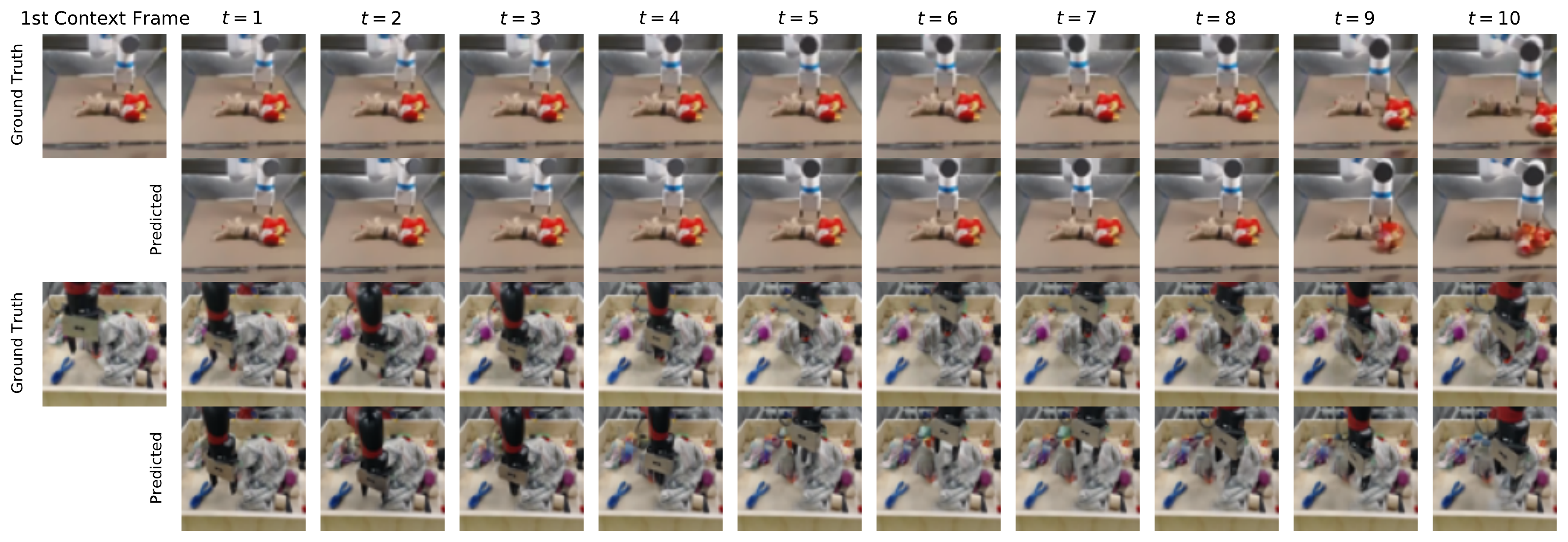}
        \caption{More detailed video from Figure~\ref{fig:robonet} which illustrates \model on action-conditioned RoboNet~\cite{dasari2019robonet}. The model is conditioned on the first two frames and is predicting the next ten frames given the future actions of the robotic arm. These figures demonstrate how the predicted movements of the arm closely follows the ground truth given that the future actions is known. The model also predicts detailed movements of the pushed objects (visible in the top example) as well as filling in the previously unseen background with some random objects (look at the object that appear behind the robotic arm in the bottom example). Also notice the wrong predictions of robots fingers in the bottom example. }
        \label{fig:robonet_full}
    \end{figure}
    
    \begin{figure}
        \includegraphics[width=\linewidth]{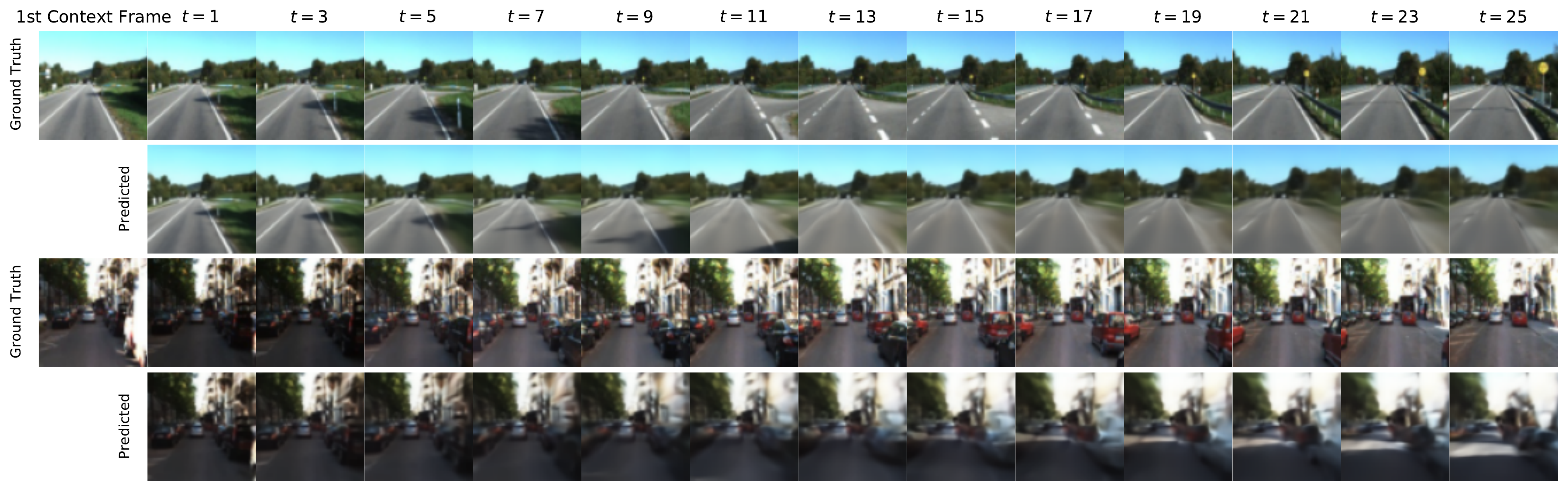}
        \caption{More detailed video from Figure~\ref{fig:kitti} which illustrates \model on KITTI dataset~\cite{geiger2013vision}. As it can be seen in this figure, the model generates high quality prediction of the future in a dynamic scene. Note how in the top example \model keeps predicting the movement of the shadow on the ground till it moves out of the frame. After that, the model keeps pushing the background closer in each frame, implying driving forward. We noticed that the quality of predictions drop substantially faster when there are more objects in the scene e.g. the driving scenes inside a city as can be seen in the bottom example. This indicates the model still fails to generalize to more complex scenes with more moving subjects.}
        \label{fig:kitti_full}
    \end{figure}
    
    \begin{figure}
        \includegraphics[width=\linewidth]{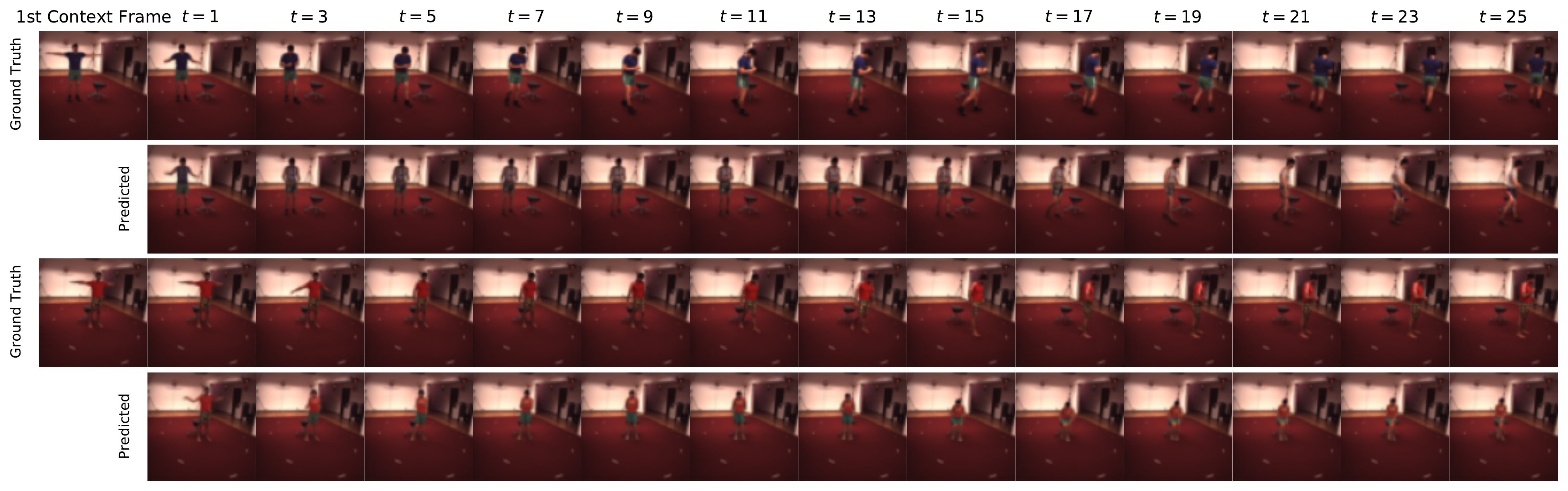}
        \caption{More detailed video from Figure~\ref{fig:humans} which illustrates \model on Human3.6M~\cite{ionescu2014human3}. This figure demonstrates extremely detailed and human-like motions predicted by \model, conditioned on the given context frames. However, on closer inspection, it can be seen that the human subject in the video is changing, from the test subject to a training subject. This is particularly evident from the cloths. This phenomena indicates that, although \model is capable of generalizing to the frames out of training distribution, however, it morphs the human subject into a familiar one from the training set and then plays the video from the memory.}
        \label{fig:humans_full}
    \end{figure}
    
    \begin{figure}
        \includegraphics[width=\linewidth]{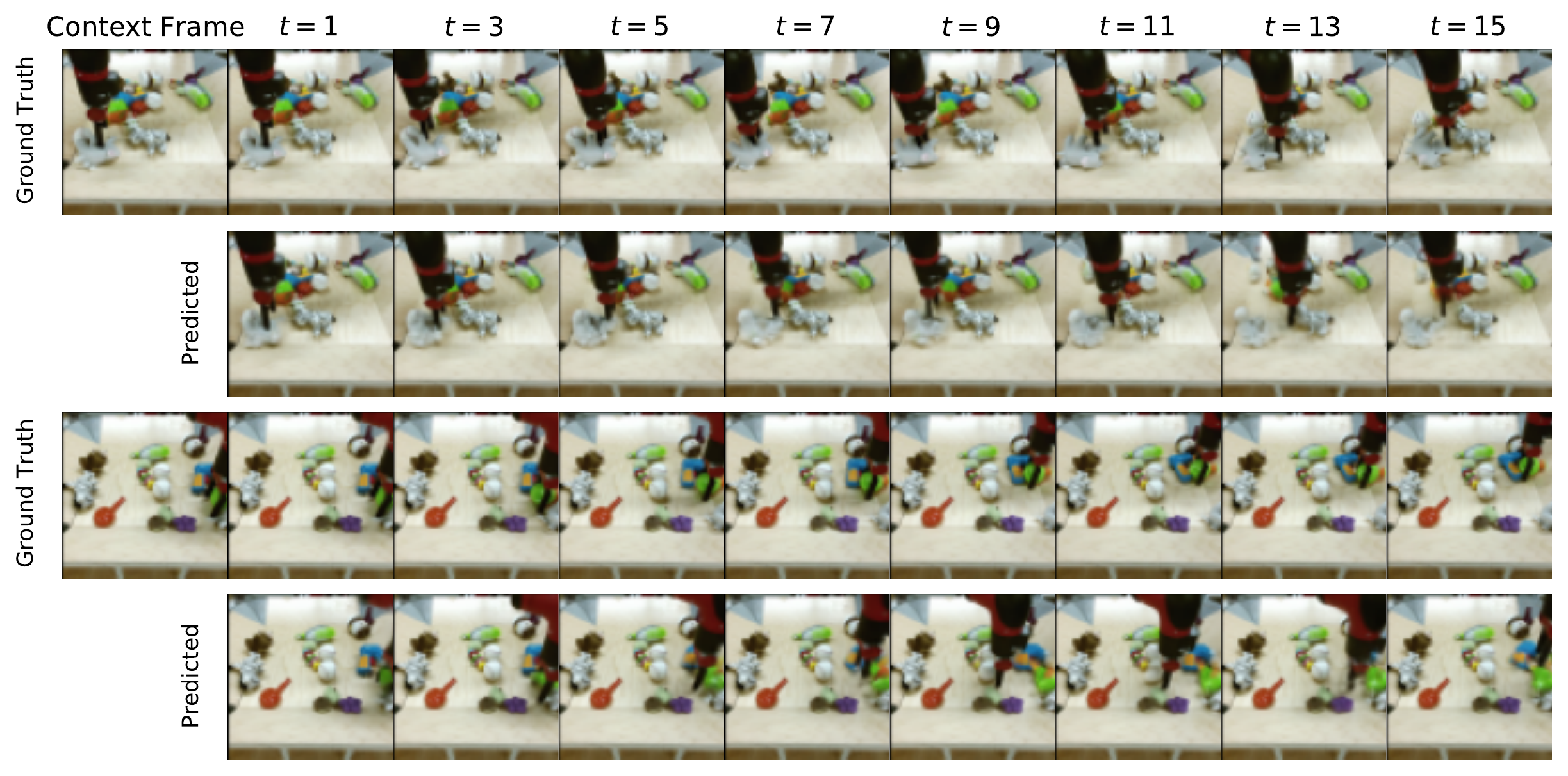}
        \caption{More detailed video from Figure~\ref{fig:bair} which illustrates \model on BAIR robot pushing dataset~\cite{2017arXiv171005268E} with no actions. The model is conditioned only on the first frame and is predicting the next 16 frames. Given that the future actions of the robotic arm is unknown, the prediction can diverge substantially from the ground truth video. However, the model predicts movements for the objects whenever the arm pushes the object in an imaginary scenario. It also fills the background with random objects.}
        \label{fig:bair_full}
    \end{figure}

\end{landscape}

\begin{figure}
  \centering
  Copy of Figure~6 from \citet{franceschi2020stochastic}
  \includegraphics[width=\textwidth]{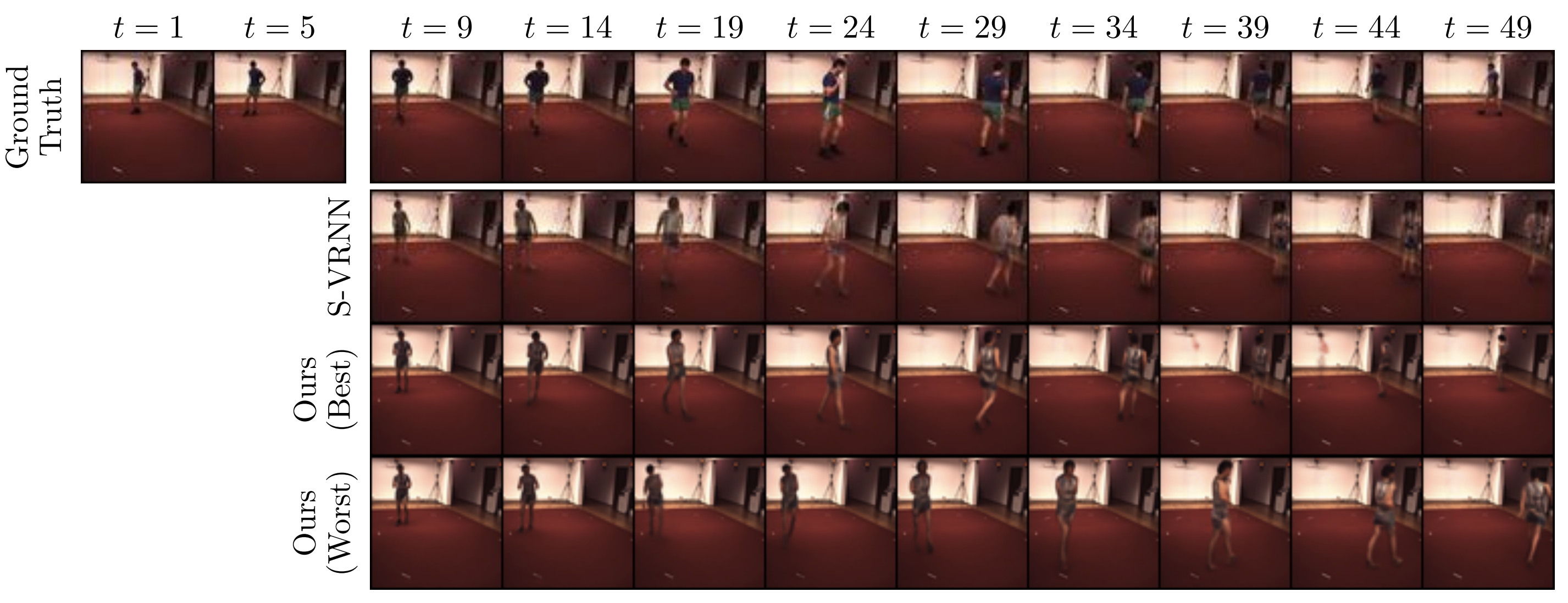}
  Copy of Figure~7 from \citet{villegas2019high}
  \includegraphics[width=\textwidth]{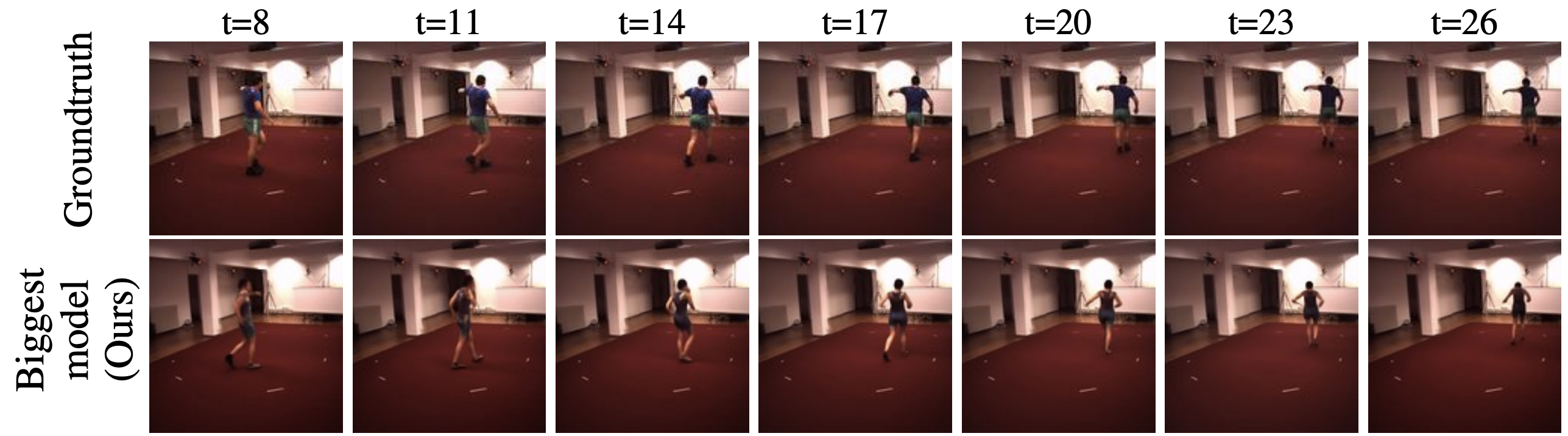}
  Copy of Figure~5 from~\citet{villegas2017learning}
  \includegraphics[width=\textwidth]{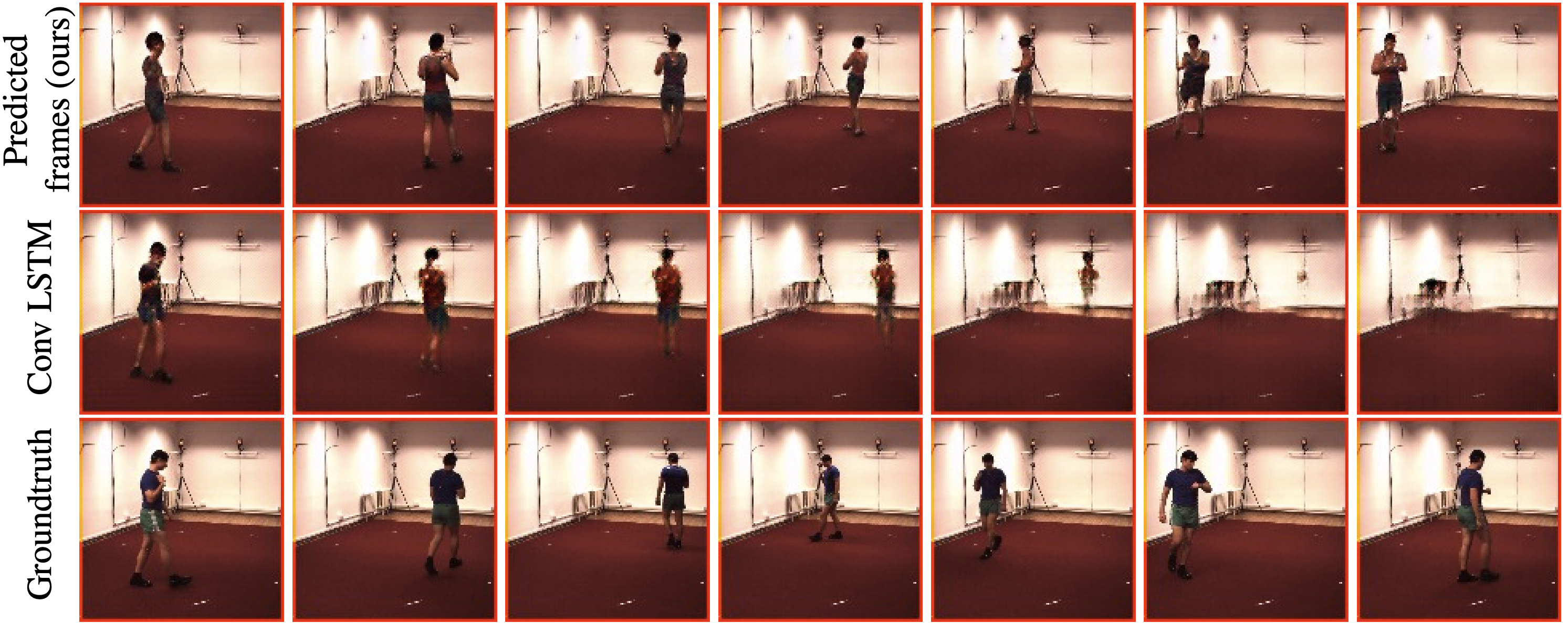}
  
  \caption{These are copies of Figure~6 from \citet{franceschi2020stochastic}, Figure~7 from \citet{villegas2019high} and Figure~5 from~\citet{villegas2017learning}. The proposed methods in these papers are also changing the human test subject into a training subject (moslt visible in the changed shirt.). This seems to be a common issue which is typically overlooked in the video prediction literature.}
  \label{fig:otherpapers}
\end{figure}

\end{document}